\documentclass[twocolumn]{article}

\usepackage{multirow}
\usepackage{colortbl}
\usepackage{subcaption}
\usepackage[table]{xcolor}
\usepackage[ruled,vlined,linesnumbered,noend]{algorithm2e}
\DontPrintSemicolon
\usepackage{amsmath}
\usepackage{amssymb}
\usepackage{bm}
\usepackage{hyperref}
\usepackage[margin=2cm]{geometry}
\usepackage{graphicx}
\usepackage[all]{nowidow}

\newcommand{\pluseq}{\mathrel{+}=}
\newcommand{\env}{\mathcal{E}}
\newcommand{\var}{\textit}  % shorthand alias for hyperparams and vars in code
\newcommand{\flsp}{p_{\text{slip}}}

\begin{document}

\title{Pittsburgh Learning Classifier Systems for Explainable Reinforcement Learning: Comparing with XCS\,\footnote{© ACM 2022. This is the author's version of the work. It is posted here for your personal use. Not for redistribution. The definitive Version of Record was published in \textit{GECCO '22: Proceedings of the Genetic and Evolutionary Computation Conference}, \url{http://dx.doi.org/10.1145/3512290.3528767}.}}

\author{Jordan T. Bishop\\
        The University of Queensland, Australia\\
        \texttt{j.bishop@uq.edu.au}
        \and
        Marcus Gallagher\\
        The University of Queensland, Australia\\
        \texttt{marcusg@uq.edu.au}
        \and
        Will N. Browne\\
        Queensland University of Technology, Australia\\
        \texttt{will.browne@qut.edu.au}}

\date{}
\maketitle

\section*{Abstract}
Interest in reinforcement learning (RL) has recently surged due to the application of deep learning techniques, but these connectionist approaches are opaque compared with symbolic systems.
Learning Classifier Systems (LCSs) are evolutionary machine learning systems that can be categorised as eXplainable AI (XAI) due to their rule-based nature.
Michigan LCSs are commonly used in RL domains as the alternative Pittsburgh systems (e.g. SAMUEL) suffer from complex algorithmic design and high computational requirements; however they can produce more compact/interpretable solutions than Michigan systems.
We aim to develop two novel Pittsburgh LCSs to address RL domains: PPL-DL and PPL-ST.
The former acts as a “zeroth-level” system, and the latter revisits SAMUEL's core Monte Carlo learning mechanism for estimating rule strength.
We compare our two Pittsburgh systems to the Michigan system XCS across deterministic and stochastic FrozenLake environments.
Results show that PPL-ST performs on-par or better than PPL-DL and outperforms XCS in the presence of high levels of environmental uncertainty.
Rulesets evolved by PPL-ST can achieve higher performance than those evolved by XCS, but in a more parsimonious and therefore more interpretable fashion, albeit with higher computational cost.
This indicates that PPL-ST is an LCS well-suited to producing explainable policies in RL domains.

\section{Introduction}
\label{sec:intro}

%\textbf{Scope, justification, hypothesis, aims, objectives.}
%Hypothesis: As grid size increases and transition stochasticity increases, both PPL variants will achieve higher mean testing performance than XCS.
In recent years, reinforcement learning (RL) has seen a surge in interest due the application of deep learning techniques to multi-step decision problems such as game playing \cite{mnih_human-level_2015} and robotics \cite{levine_end--end_2016}.
However, the solutions produced by such connectionist systems are inherently opaque and cannot be easily interpreted by humans, in contrast to symbolic systems.
Learning classifier systems (LCSs) are a family of evolutionary machine learning systems that have been studied in RL domains for many years.
A key attraction of these systems is that they represent their solutions using sets of symbolic \textit{IF-THEN} rules, and so are capable of producing human interpretable solutions.
This places LCSs under the umbrella of eXplainable Artificial Intelligence (XAI) \cite{adadi_peeking_2018}.
Given the increasing interest in XAI, in this work we choose to focus our attention solely on LCSs and defer a comparison with other approaches.

Within the LCS family, two main types of systems are identified: Michigan and Pittsburgh \cite{urbanowicz_introduction_2017}; both employing population-based methods to learn.
In Michigan systems, an individual is a single rule, and all individuals in the population form the ruleset in ensemble.
Pittsburgh systems employ the traditional evolutionary approach where each individual in the population is an entire solution (i.e. ruleset).
Most research applying LCSs to RL involves the use of Michigan systems, the most well-studied being XCS \cite{wilson_classifier_1995} and its variants.
Pittsburgh LCSs have not been well-studied in RL domains, one possible reason being their need for large amounts of computing power; for example evaluating the fitness of a single ruleset could involve simulating an entire game playing round.
However, this drawback is increasingly less significant with the continuous expansion of computing power.

The most well-studied Pittsburgh system for RL is SAMUEL \cite{grefenstette_lamarckian_1995}, which makes use of many algorithmic features in pursuit of increasing its performance.
However, this significantly increases the complexity of its design; impeding understanding of how it operates and making implementation difficult.
This situation of over-complexity in SAMUEL is reminiscent of the designs of the first Michigan systems: CS-1 and its descendants, which underwent a simplification process in Wilson's work on the zeroth-level system ZCS \cite{wilson_zcs_1994}, and later extension to produce XCS.
No such simplification process has yet been applied to SAMUEL for RL domains.
Therefore, in this work we develop two variants of a Pittsburgh LCS designed for RL that we dub Pittsburgh Policy Learner (PPL).
The first, PPL-DL (PPL-\textbf{D}ecision\textbf{L}ist, Section~\ref{1sec:ppldl}), is designed to be a simple as possible (a zeroth-level system), acting as a baseline.
The second, PPL-ST (PPL-\textbf{ST}rength, Section~\ref{1sec:pplst}), is more complex, and revisits the core Monte Carlo (MC) rule strength learning mechanism of SAMUEL (in a simpler manner than SAMUEL itself) in an effort to re-evaluate the usefulness of this mechanism.
The first objective of this work is the construction of these two PPL variants.

We wish to understand how our Pittsburgh systems compare to existing Michigan approaches in RL domains, similar to previous work done comparing Pittsburgh and Michigan systems in supervised learning (SL) \cite{bernado_xcs_2002,kovacs_data_2007}.
Thus, the second objective of this work is to compare both PPL variants to the Michigan system XCS across a suite of RL tasks.
FrozenLake (FL) environments differ from more traditional Woods or Maze environments in that they feature varying levels of transition stochasticity in addition to sparse reward and long action chains (the latter two features can cause difficulty for XCS, see e.g. \cite{barry_stability_2002,stein_xcs_2020}), making them a challenging set of benchmarks to use as our problem suite.
Our comparisons aim to make qualitative comments about the learning behaviours of the different systems, in addition to quantitatively determining which systems are better suited to certain problems.

We form a hypothesis about this series of comparisons that we term the \textit{Pittsburgh Strength Hypothesis} (PSH), which states:
``In FL environments, as both the length of action chains and amount of environmental uncertainty increase, PPL-ST will consistently achieve higher performance than both i) PPL-DL and ii) XCS.''
Our reasoning is that we expect PPL-ST is able to identify robust and high-performing sub-chains of rules that can be combined to form high-performing policies, compared to the other systems that struggle with this as the problems increase in difficulty.
Lastly, it transpires that fairly comparing Pittsburgh and Michigan systems in RL domains is not as straightforward as it is in SL domains.
Thus, the third objective of this work is to construct an experimental framework in which the two types of systems can be fairly compared, setting the ground work for similar future comparisons.

The remainder of this work proceeds as follows: Section~\ref{sec:related} discusses related work, Section~\ref{sec:background} details background relevant to the novel methods and benchmarks used, Section~\ref{sec:pitt} gives design considerations and descriptions of both PPL systems, Section~\ref{sec:comp_method} constructs the fair comparison methodology used in Section~\ref{sec:results}, which shows results of applying our learning methods to the FL environment suite; after which results are discussed in Section~\ref{sec:discussion} and conclusions are drawn in Section~\ref{sec:conclusion}.
%The remainder of this work proceeds as follows: in Section~\ref{sec:related} we discuss related work that compares Michigan and Pittsburgh systems in SL. Section~\ref{sec:background} details background relevant to the novel methods and benchmark approaches used.
%Next, design considerations for PPL-DL and PPL-ST, along with system descriptions, are presented in Section~\ref{sec:pitt}.
%Section~\ref{sec:comp_method} constructs the fair comparison methodology that is used in Section~\ref{sec:results}, where we show results of applying our learning methods to the FL environment suite, after which Section~\ref{sec:discussion} is used to discuss these results and address the PSH.
%Finally, the work is concluded in Section~\ref{sec:conclusion}, detailing contributions and recommendations for future work.

\section{Related Work}
\label{sec:related}
SAMUEL, which we describe in detail in Section~\ref{1sec:samuel}, is the most recent Pittsburgh system (Grefenstette 1995, \cite{grefenstette_lamarckian_1995}) to directly address RL domains, however there have been many Pittsburgh systems described in the literature for SL domains.
One notable pair of such systems is GAssist \cite{bacardit2004pittsburgh}, which was designed for general-purpose data mining tasks, and its successor BioHEL \cite{bacardit_automated_2007}, that was introduced in the context of bioinformatics.
Both are heavily acclimated to the paradigm of SL, and use an incremental learning approach, which could possibly be adapted to address RL, but needs further investigation.
%    GAssist introduced a mechanism to adapt the discretisation of decision intervals, allowing said boundaries to be fine-tuned in a more directed way rather than relying solely on crossover and mutation.
%    BioHEL further introduced a specialised knowledge representation that is useful for effectively dealing with datasets containing mixed discrete-continuous attributes.
GAssist and BioHEL use a genetic algorithm (GA) to evolve a population of variable-length rulesets; these being considered ordered and including a default action; thus embodying an \textit{IF-THEN-ELSE} \textit{decision list} structure.
To cope with the problem of bloat, the GA fitness function utilises the Minimum Description Length (MDL) principle to balance accuracy and complexity of rulesets \cite{bacardit_bloat_2007}.
The decision list ruleset structure is directly transferable to RL (see Section~\ref{1sec:ppldl}), but much investigation is required to determine if MDL-based fitness is applicable.
Another Pittsburgh system for SL is GALE \cite{llora_knowledge-independent_2001}, which utilises a fine-grained parallel evolutionary algorithm to evolve solutions to SL problems.
It is capable of evolving competing solution representations (e.g. decision trees and \textit{IF-THEN} rules), and selecting the one which performs best.
Due to its computational requirements, this architecture is unlikely to scale well to RL domains.
We emphasise that GAssist, BioHEL, and GALE are all specifically designed for SL domains, and therefore it is not sensible to apply them to RL domains without substantial modifications.
This motivates the construction of our PPL systems for RL domains via a ``first principles'', simplicity-driven approach.

There have been various experimental studies conducted that compare the aforementioned Pittsburgh systems to Michigan systems in SL domains, e.g. \cite{bernado_xcs_2002,kovacs_data_2007,fernandez_genetics-based_2010}.
However, the methodologies employed in these studies are not directly transferable to RL, as we discuss further in Section~\ref{sec:comp_method}.
When surveying the literature, we did not encounter any comparisons in RL domains between Pittsburgh Michigan systems.
Therefore, as far as we are aware, such comparisons remain an open area of investigation.
%There have been various experimental studies conducted that compare these  Pittsburgh systems to Michigan systems in SL domains.
%For example, in [bernado et al.] XCS is compared to GALE across fifteen classification problems.
%Both systems were evaluated based on their predictive accuracy attained via ten fold cross-validation --- a standard comparison methodology for SL.
%Results indicated that there were no significant differences between XCS and GALE in terms of prediction accuracy, but that both systems evolved differing rulesets (in terms of structure and size).

%Another example is the work of Bacardit and Butz in [bacardit and butz], where XCS is compared with GAssist, again across several classification domains.
%Because XCS and GAssist operate in quite different manners, the comparison methodology involved allowing each system to learn for a sufficient amount of time, then comparing predictive results.
%It was observed that XCS had a tendency to overfit the training data, whereas GAssist tended to avoid this behaviour.
%On the other hand, GAssist had some issues when the number of classes were large, attributed to the difficulty of its larger search space.
%As in the previous study, the overall conclusion was that both systems were suitable for the types of domains investigated, and neither system was clearly superior to the other.

\section{Background}
\label{sec:background}
\subsection{Reinforcement Learning}
\label{1sec:rl}
RL involves an agent interacting with an environment $\env$, which in this work is episodic, being characterised as a Markov Decision Process (MDP) formed by the tuple $(S, A, P, R, \gamma, t_{\text{max}})$ where: $S$ is the state space, $A$ is the action space, $P(s' \vert s, a)$ is the transition function, $R(s, a, s')$ is the reward function, $\gamma \in [0,1]$ is the discount factor, and $t_{\text{max}}$ is the maximum episode length \cite{sutton_reinforcement_2018}.
We further define $S_I \subseteq S$ to be the set of initial states that the environment can inhabit at the start of an episode, and $z$ to be a \textit{sequence} of initial states (duplicates allowed) drawn from $S_I$.
Why we define $z$ as a sequence will become clear in Section~\ref{sec:results}.

The task of the agent is to develop a \textit{policy}, $\pi: S \rightarrow A$, that maximises the expected amount of reward it receives from the environment in the long-term. 
For episodic environments, this concept is quantified by the \textit{return} $G$, which is the cumulative discounted sum of rewards obtained from a particular starting state $\zeta \in z$, calculated as: $G(\zeta) = \sum_{t=0}^{t_{\text{max}}}{\gamma^t \cdot r_t}\, \vert\, s_0 = \zeta$. Here $r_t$ denotes the reward obtained at time step $t$, and $s_0$ denotes the initial state of the episode.
Maximising the expected return over the sequence of initial states $z$, which we term the \textit{performance} of a policy, can therefore be calculated as:
\begin{equation}
    \label{eqn:perf}
    \text{performance} = \frac{1}{\vert z \vert}\sum_{\zeta \in z} G(\zeta)
\end{equation}
where $G(\zeta)$ is calculated via performing a \textit{rollout} of the policy in the environment.
An additional concept is that of a value function, specifically an \textit{action-value} or \textit{Q-function}, $Q: S \times A \rightarrow \mathbb{R}$.
This function gives the utility of each state-action pair, measured as the amount of expected discounted cumulative reward obtainable by performing action $a$ in state $s$ and then following a particular policy $\pi$ thereafter, i.e. the return achievable from $s$ via following $\pi$.
$Q^*$ is used to denote the special case where the policy being followed is the \textit{optimal policy}, meaning $Q^*$ maximises utility across the entirety of $S \times A$.

\subsection{SAMUEL}
\label{1sec:samuel}
SAMUEL was originally described by Grefenstette et al. in \cite{grefenstette_learning_1990}, and can be summarised as a Pittsburgh LCS that uses a GA to evolve variable-length, non-ordered \& overlapping rulesets that the authors term``strategies''.
We use the term SAMUEL-v1 to refer to this initial version of the system.
Although the authors do not explicitly reference RL, it is clear that the system is evolving policies for an RL-esque problem; it learns chains of actions from delayed rewards.
An interesting aspect of SAMUEL is that it assigns credit to singular rules within a ruleset in addition to assigning credit to rulesets as a whole (i.e. the standard concept of genetic fitness).

To achieve the former, a learning mechanism is used to estimate the \textit{strength} (utility) of rules within a ruleset, using a basic MC form of RL.
Specifically, between each iteration credit is apportioned to rules on an episodic basis: the ruleset under evaluation is used to generate a return in the environment, which the authors label as the payoff $r$ (we label this as $P$ to avoid confusion with individual rewards).
While doing this, rules that participate in action sets are marked as \textit{active}.
Each rule maintains an estimate of i) the mean $\mu$, and ii) the variance $v$ of the payoff it receives when it is active.
Upon receipt of $P$, all rules that were active during the episode have their estimates updated as:
\begin{equation}
    \label{eqn:samuel_mu}
    \mu = (1-c) \cdot \mu + c \cdot P 
\end{equation}
\begin{equation}
    \label{eqn:samuel_v}
    v = (1-c) \cdot v + c \cdot (\mu - P)^2
\end{equation}
where $c$ is the learning rate.

Notice that \textit{all active rules} are updated to estimate the same level of payoff, as the specific times at which they are active within the episode are not recorded.
Consider now a situation where the minority of $P$ is received at the start of an episode, and the remaining majority of $P$ is received at the end of the episode.
This scheme will assign the same amount of credit to two separate rules that fire at the beginning and end of the episode.
This is an odd design decision, as it means SAMUEL cannot differentiate between different levels of payoff within a chain of rules.
Furthermore, this mechanism does not use any form of reward discounting.

Notice also that the mean level of payoff predicted by each rule is a scalar value.
This implies that the system is using an \textit{egocentric} (position relative to environmental surroundings) representation.
This is similar to the initial descriptions of XCS, e.g. XCS with ternary rule representation and scalar predictions employed in maze-like environments \cite{wilson_classifier_1995}.
In this scenario, each rule is considered to be a certain ``distance'' away from an amount of payoff, and so estimating payoff with a scalar makes sense.
When Wilson introduced the concept of computed prediction in XCSF \cite{wilson_classifiers_2001}, it enabled XCS to predict differing levels of payoff for a single rule, because an \textit{allocentric} (absolute position within the environment) representation was used.
It is worth keeping both the lack of reward discounting and egocentric nature of SAMUEL in mind as we move forward in this work.

Returning to rule strength, SAMUEL uses the estimates $\mu$ and $v$ to calculate rule strength as a lower bound on the expected level of payoff received:
\begin{equation}
    \label{eqn:samuel_strength}
    \text{strength} = \mu - \sqrt{v}
\end{equation}
\vfill\null
This has the effect of considering rules to be of higher quality if they lead to high levels of payoff and also do so consistently.
SAMUEL uses rule strength as a basis for conflict resolution within a ruleset, giving preference to actions that are advocated by rules with high strength.

The authors tested SAMUEL-v1 in a simulated flight environment where the system was tasked with developing policies for missile evasion.
Performance of the system was measured as success rate of evasion over multiple episodes.
%A key concern was investigating discrepancies between the environments the system was trained and tested on --- something we do not consider in this work but which is nonetheless valuable.
Environmental setups incorporated variation in training/testing conditions, initial episode conditions, and the inclusion or absence of sensory noise.
Results showed that SAMUEL-v1 had the highest success rate ($\ge 80\%$) when training and testing conditions were identical, including when initial conditions were variable and sensory noise was introduced.
However, it was also observed that system performance was fairly robust even in adverse training conditions (e.g. training on noise-free environment, testing on noisy environment).
This led the authors to conclude that SAMUEL-v1 was adept at exploiting regularities in the environment.

Grefenstette later investigated extending SAMUEL with a set of Lamarckian learning operators \cite{grefenstette_lamarckian_1995}.
We dub this version of the system SAMUEL-v2.
The intention of this work was to determine if said operators were effective at improving overall system performance via an exploitative search mechanism used to improve specific rules within rulesets.
However, SAMUEL-v2 was only compared with SAMUEL-v1 on a single environment (a more difficult evasion environment than from \cite{grefenstette_learning_1990}), thus not providing strong evidence that the introduced operators were \textit{generally} helpful across multiple domains.
Overall, a conclusion that can be drawn from both pieces of work is that if Pittsburgh LCSs applied to RL were to be revisited, it would be wiser to firstly reconsider SAMUEL-v1 rather than continuing with SAMUEL-v2.

\subsection{XCS}
\label{1sec:xcs}
In RL domains, XCS can be summarised as an \textit{accuracy-based} LCS that uses a population of classifiers (individual rules) to collectively learn a \textit{complete action map} of the environmental action-value function.
It does so in an \textit{online} fashion, using a hybrid of \textit{temporal difference} (Q-learning like) reinforcement learning and a steady-state GA to produce accurate, maximally general rules.
XCS should ideally learn about every possible state-action pair (the entirety of $S \times A$) in order to properly disambiguate different actions.
Since XCS is learning an approximation of the optimal action-value function $Q^*$ from data, we denote its learned approximation as $\hat{Q}$.
To construct XCS's behavioural policy, $\hat{Q}$ is greedily queried over $s \in S$.

$\hat{Q}$ is computed as a fitness-weighted average of classifier predictions in an action set $[A]$: $\hat{Q}(s, a) = \frac{\sum_{cl \in [A]} cl.p(s)\; \cdot\; cl.f}{\sum_{cl \in [A]}cl.f}$, where $cl$ denotes a classifier, $cl.f$ denotes classifier fitness and $cl.p(s)$ denotes the (computed) prediction of a classifier given state $s$.
Since classifier fitness is \textit{based on accuracy}, classifiers that have learnt accurate predictions of payoff have higher fitness and thus contribute more to the overall system prediction.
%The Michigan system that we employ in this work is xcs; i.e. XCS with an \textit{allocentric} rule representation and computed prediction (see Section~\ref{sec:results} for details).
%We will simply refer to this system as XCS for the remainder of this work.
\newpage
\section{Pittsburgh Systems}
\label{sec:pitt}
\subsection{Design Considerations}
\label{1sec:pitt_design_considerations}
The first objective of this work is to construct two variants of the Pittsburgh LCS PPL.
An initial design consideration for PPL is, how should rulesets be structured?
This has two components: 1) how does \textit{inference} (the process of selecting an action given an input state) operate?, and 2) are the rulesets fixed-length or variable-length?
Addressing the first component, in \cite{fernandez_genetics-based_2010}, possible ruleset inference structures identified are: i) non-ordered \& overlapping rulesets, ii) non-ordered \& non-overlapping rulesets, iii) ordered rule sets (decision lists).
In the first case, some kind of conflict resolution mechanism is required. In the second (and most restrictive) case, there must be some mechanism to prevent rules from overlapping by dividing the input space into non-overlapping regions, which is is likely possible but requires additional algorithmic complexity.
The third case is the simplest, as the ordering itself performs conflict resolution.
Addressing the second component, it is known that variable-length rulesets can be associated with the problem of bloat \cite{langdon_fitness_1998}, necessitating a mitigation mechanism.
Fixed-length rulesets do not suffer from bloat, but are obviously more restrictive.
Despite this, we select fixed-length rulesets due to their simplicity.
A second design consideration is: what type of evolutionary process should be used in PPL?
To keep things simple, we use a canonical GA as a basis.
The population size ($\var{popSize}$) for this GA is required to be even, which will become important in Section~\ref{sec:comp_method}.

\subsection{Design Commonalities}
\label{1sec:pitt_design_commonalities}
Before continuing, some commonalities across both PPL variants need to be discussed.
For both systems, during inference there is the possibility that no particular rule in a ruleset covers a given input state (due to the fact we do not include a covering operator or use a default action), and thus the system is unable to select any action.
In this case, the system returns a null action --- the consequences of this are explained in Section~\ref{sec:results}.
For rule conditions we use hyperrectangles, encoded using Unordered Bound Representation (UBR) (see \cite{stone_for_2003} for details).
Assuming the dimensionality (number of features) of $S$ is $d$, conditions are genotypically encoded as $2d$ alleles, one pair for each dimension.
The genotype of each rule also specifies a single action allele $a \in A$.
Let $\var{idvSize}$ represent the (fixed-length) ruleset size for PPL.
At the start of learning, each individual in the population is initialised to contain $\var{idvSize}$ rules whose alleles are initialised to random values.
For selection we use tournament selection with parameter $\var{tournSize}$.
The fitness of an individual is equal to the performance of its ruleset (application of Equation~\ref{eqn:perf}).

Crossover is applied with probability $\var{pCross}$, and is different for both PPL variants (differences detailed shortly), however mutation is the same: each allele in each rule is mutated independently with probability $\var{pMut}$.
For condition alleles, since in this work we address problems with integer state components (see Section~\ref{sec:results}), mutation noise is drawn from a geometric distribution: $\pm\text{Geo}(p)$, supported on integers $\ge 1$. $p$ is calculated on a per state dimension basis such that $99\%$ of the distribution mass lies within half of the dimension width $w$, i.e.: $[1, \lfloor \frac{w}{2} \rfloor]$.
For action alleles, the represented action $a$ is mutated to a random action in $A - \{a\}$.

\subsection{PPL-DL}
\label{1sec:ppldl}
For our first PPL variant we use a decision list ruleset representation, and name the system PPL-DL.
In PPL-DL, each rule is composed of only a condition and an action.
The inference process is to simply: return the action of the first rule in the decision list that matches the input state, and if no matches are found, return a null action (no default action is employed).
Crossover in PPL-DL is uniform crossover applied on a per allele basis, i.e. crossover points can occur \textit{within} rule boundaries.
PPL-DL can be summarised as a Pittsburgh LCS that uses a population of rulesets to perform \textit{direct policy search} on the environmental performance function, learning in an \textit{offline} manner.
It is an \textit{implicitly strength-based} system, since rules do not keep explicit estimates of their strength but are rather favoured for inclusion in future rulesets if they lead to better performance.
Each ruleset forms a \textit{partial action map} \cite{kovacs_strength_2004} across $S \times A$.

\subsection{PPL-ST}
\label{1sec:pplst}
The second variant of PPL is dubbed PPL-ST, due to its usage of rule strength.
Like SAMUEL, PPL-ST uses a non-ordered \& overlapping ruleset representation, and because of this a conflict resolution mechanism is required.
The approach we have taken is inspired by the conflict resolution of SAMUEL, as detailed in Section~\ref{1sec:samuel}.
Recall that we described SAMUEL as an egocentric system, because of the fact that individual rules maintain scalar mean payoff estimates.
We wish to use the same notion of strength from SAMUEL in PPL-ST, but to do so in an allocentric manner, because the problems we address in this work have allocentric state representations (see Section~\ref{sec:results}).
Thus, we must perform a shift to allocentrism in PPL-ST that is akin to that done when moving from XCS to XCSF.
Therefore, rules in PPL-ST contain the following components: i) a condition, ii) an action, iii) a prediction weight vector $\vec{w}$, and iv) an estimate of the prediction variance $\var{var}$.

Each rule's weight vector allows its payoff prediction to be computed across the region it covers; the simplest possible scheme being linear computed prediction.
Therefore a hyperparameter $x_0$ (as in XCSF) is introduced, with the augmented state vector $x$ being produced by prepending $x_0$ to $s$.
Then, a rule's estimate of the payoff is computed via a dot product between its weight vector $\vec{w}$ and $x$; represented by $\var{rule.pred}(x)$.
The strength of a rule is now computed as: $\text{strength} = \var{rule.pred}(x) - \sqrt{\var{rule.var}}$ (c.f. Equation~\ref{eqn:samuel_strength}).
The inference process for PPL-ST involves a ``double max'': given an input state, each action is advocated with the maximum strength of any rule in its corresponding action set. Then, the action with the maximum advocated strength is selected.
Like for PPL-DL, if no rules match the input state, a null action is returned (again see Section~\ref{sec:results} for implications).

Rule payoff predictions must be learnt via experience in the environment.
To accomplish this, a procedure \texttt{reinforceRules}, described in Algorithm~\ref{alg:pplst_reinforceRules}, is used\footnote{Note the similarities between this procedure and the policy gradient algorithm REINFORCE from RL (consult \cite{sutton_reinforcement_2018}); however consider that our procedure updates heterogeneous rules that constitute a policy rather than updating homogeneous policy parameters used by e.g. a neural network.}.
This procedure is conducted for each individual at each generation, immediately before fitness evaluation; operating as an inner learning loop within the outer evolutionary loop.
\begin{algorithm}
	\KwIn{Environment $\env$, individual to reinforce $\var{idv}$}
	\For{$\var{numReinfRollouts}$ \textup{times}}{
		$\tau = \texttt{genTrajectory}(\env, \var{idv})$\;\label{alg:pplst_reinforceRules_tau}
		$T = $ num. entries in $\tau$\;
		$\var{rSum} = 0$\;
		\For{$i = T-1 $ \textup{down to} $0$}{
			$(s, a, [A], r) = i^{\,th} $ entry of $\tau$ (0 indexed)\;
			$\var{rSum} \pluseq r$\;
			$j = T-1-i$\;\label{alg:pplst_reinforceRules_j}
			$P = \gamma^{\,j} \cdot \var{rSum}$\;\label{alg:pplst_reinforceRules_P}
			$\texttt{updateActionSet}([A], P, s)$\;
		}
	}
	\caption{PPL-ST: $\texttt{reinforceRules}$}
	\label{alg:pplst_reinforceRules}
\end{algorithm}
The rules in a given ruleset are reinforced in an MC fashion by, $\var{numReinfRollouts}$ times, generating a \textit{trajectory} $\tau$ using the ruleset (line~\ref{alg:pplst_reinforceRules_tau}), then iterating backwards over the elements of $\tau$ to incrementally calculate and reinforce payoffs for rules that were part of action sets.
The initial state used to generate $\tau$ is selected randomly from $S_I$.
Each element of $\tau$ is a 4-tuple $(s, a, [A], r)$: storing the state, action, action set, and reward for a time step.

For each backwards iteration over $\tau$, the payoff $P$ is calculated as the discounted sum of rewards encountered so far (discounting progressively applied from the end of $\tau$ --- refer to lines ~\ref{alg:pplst_reinforceRules_j} \& \ref{alg:pplst_reinforceRules_P} where $j$ is the number of steps from the end of $\tau$).
Crucially, and in contrast to the rule-level credit assignment performed in SAMUEL (c.f. Section~\ref{1sec:samuel}), PPL-ST is capable of reinforcing differing levels of payoff within a trajectory, since discounting is used.
Given $P$, all rules in $[A]$ have their payoff estimates updated via the \texttt{updateActionSet} procedure, shown in Algorithm~\ref{alg:pplst_updateActionSet}.
\begin{algorithm}
	\KwIn{Action set $[A]$, payoff $P$ to regress towards, applicable state $s$}
	$x = $ augment state vector $s$ by prepending scalar $x_0$\;
	\For{$\var{rule} \in [A]$}{
		$\epsilon = P - \var{rule.pred}(x)$\;\label{alg:pplst_updateActionSet_e}
		$\var{rule.}\vec{w} \pluseq \frac{\eta}{\Vert x \Vert^2_2} \cdot \epsilon \cdot x$\;\label{alg:pplst_updateActionSet_w}
		$\var{rule.var} = (1 - \eta) \cdot \var{rule.var} + \eta \cdot \left(\var{rule.pred}(x) - P\right)^2$\;\label{alg:pplst_updateActionSet_var}
	}
	\caption{PPL-ST: $\texttt{updateActionSet}$}
	\label{alg:pplst_updateActionSet}
\end{algorithm}

This procedure utilises the Normalised Least Mean Squares algorithm to update rule weight vectors (lines \ref{alg:pplst_updateActionSet_e} \& \ref{alg:pplst_updateActionSet_w}).
This is the same approach used in the initial description of XCSF \cite{wilson_classifiers_2001}, selected due to its simplicity compared to other algorithms (e.g. Recursive Least Squares).
Finally, on line~\ref{alg:pplst_updateActionSet_var} the payoff variance estimate of a rule is updated; in a similar fashion to Equation~\ref{eqn:samuel_v}.
A learning rate $\eta$ is required to perform both parameter updates, a typical value for this is $0.1$.
Crossover in PPL-ST is again uniform crossover, but adopts the same approach taken in SAMUEL, which is to only allow crossover points to occur between rule boundaries, so that entire rules are exchanged.
%This is the same approach taken in SAMUEL, and we do this to reduce the amount of disruption caused to the alleles of rules.
%This means that the only mechanism of allelic alteration is mutation; which will usually only make small changes.
%We suspect, but do not know for certain, that this approach meshes well with the rule reinforcement mechanism; if rules are altered drastically with too high a frequency it may be difficult for said mechanism to preserve and build upon the learning it has already performed.
%Perhaps this is an area for future inquiry.
PPL-ST can be summarised as a Pittsburgh LCS that uses a population of rulesets to perform a hybrid of \textit{direct policy search} and \textit{Monte Carlo reinforcement learning}.
The system is \textit{explicitly strength-based}, as rules within rulesets learn estimates of payoff and payoff variance, which are used to calculate rule strength, on which conflict resolution is based.
Like PPL-DL, rulesets in PPL-ST represent a partial action map over $S \times A$.

\section{Comparison Methodology}
\label{sec:comp_method}
%Now that we have described our three learning systems, we can detail our methodology used to compare the systems; the overarching goal in this endeavour being fairness of comparison.
%First, we must discuss how the different systems learn.
%Recall that in Section~\ref{1sec:rl} we stipulated the task of an agent in RL is to maximise performance of its policy (maximise Equation~\ref{eqn:perf}), measured over the initial state sequence $z$.
Within the MDP of an RL environment there are two fundamental units of time: a time step, and an episode (a variable-length sequence of time steps).
XCS is capable of being trained in either of these two units, however PPL is only capable of being trained in the unit of episodes: with a single generation using a fixed number of episodes.
Furthermore, depending on the policies PPL has constructed, it cannot be known a priori exactly how many time steps will be consumed in each episode.
Because of this, PPL and XCS cannot be fairly compared in units of time steps, but what about in units of episodes?
Considering PPL-ST, the number of episodes used by each individual at each generation is equal to $\var{numReinfRollouts}$ (application of Algorithm~\ref{alg:pplst_reinforceRules}), plus an additional $\vert z \vert$ episodes to compute fitness (application of Equation~\ref{eqn:perf}).
Thus, for an entire generation, the number of episodes used is:
\begin{equation}
    \label{eqn:pplst_epsPerGen}
    \var{episodesPerGen} = \var{popSize}\ (\var{numReinfRollouts} + \vert z \vert)
\end{equation}
Note that this is an upper bound compared to PPL-DL, which lacks the rule reinforcement procedure; only requiring $\vert z \vert$ episodes for each individual.
Consider a case where $\var{popSize}{=}500$, $\vert z \vert{=}50$, and\\$\var{numReinfRollouts}{=}10$, which as per Section~\ref{sec:results} are realistic numbers.
Applying Equation~\ref{eqn:pplst_epsPerGen} yields: $\var{episodesPerGen} = 500\ (10 + 50) = 30{,}000$.
Accrued over multiple generations, it quickly becomes infeasible to run a Michigan system for an equivalent number of episodes.
Therefore using episodes as a time unit for comparison is impractical.

What we actually care about is comparing the three systems at some consistent step size of ``evolutionary progress'', regardless of how much computation is done between these steps.
Because XCS, PPL-DL, and PPL-ST all use a GA for their evolutionary algorithm, it is possible to compare them in terms of \textit{number of GA invocations}.
What constitutes a GA invocation in XCS is clear: breeding two children from two parents in a particular action set, the frequency of this controlled by the parameter $\theta_{\text{GA}}$.
For PPL, we define a GA invocation to be a single \textit{breeding round}. Since PPL's $\var{popSize}$ is even, there are $\frac{popSize}{2}$ breeding rounds in each generation.
This measure is also constant for both PPL variants: even though PPL-ST performs more computation between breeding rounds in its rule reinforcement procedure.
Given this, we define an \textit{epoch} to be the \textit{number of GA invocations in a single generation of PPL}.

Therefore, our comparison methodology is to run all three systems for some number of epochs, calculating their performance at each epoch: this is the ``testing'' procedure.
For PPL, this simply involves running the algorithm for an entire generation (epoch) at a time, its performance at a particular epoch being equal to the maximum performance of any ruleset in the population.
The training and testing procedures for PPL are identical, as the fitness function is equal to policy performance.
When computing performance for PPL, there is a possibility that a given ruleset does not cover a particular state, and thus returns a null action (see Section~\ref{1sec:pitt_design_commonalities}).
In this scenario, we assign PPL a performance value of zero, which is a lower bound across all environments.

For XCS, training involves running the algorithm in single time step increments, and monitoring how many GA invocations have been performed after each time step.
When the desired number of GA invocations (an epoch) is reached, the current ruleset (entire population of rules) undergoes performance testing.
Note that the training procedure for XCS is \textit{not} the same as the testing procedure.
As is usually done in the literature, during training XCS alternates between \textit{explore} and \textit{exploit} modes; choosing a mode randomly at the start of an episode.
In explore mode, we set XCS to use an $\epsilon-$greedy policy, with fixed $\epsilon{=}0.5$.
In exploit mode, XCS always selects actions greedily.
The other nuance of XCS training is that we employ the \textit{teletransportation} meta-exploration technique \cite{lanzi_analysis_1999}, which involves setting the initial state of the environment to a state randomly selected from $S_I$ at the beginning of each episode.
Note that, since $z$ is composed from enumeration(s) of $S_I$ (see Section~\ref{sec:results}), we are sampling all possible initial testing states at random for XCS during training.

\section{Results}
\label{sec:results}
\begin{figure}
    \centering
    \includegraphics[width=\columnwidth]{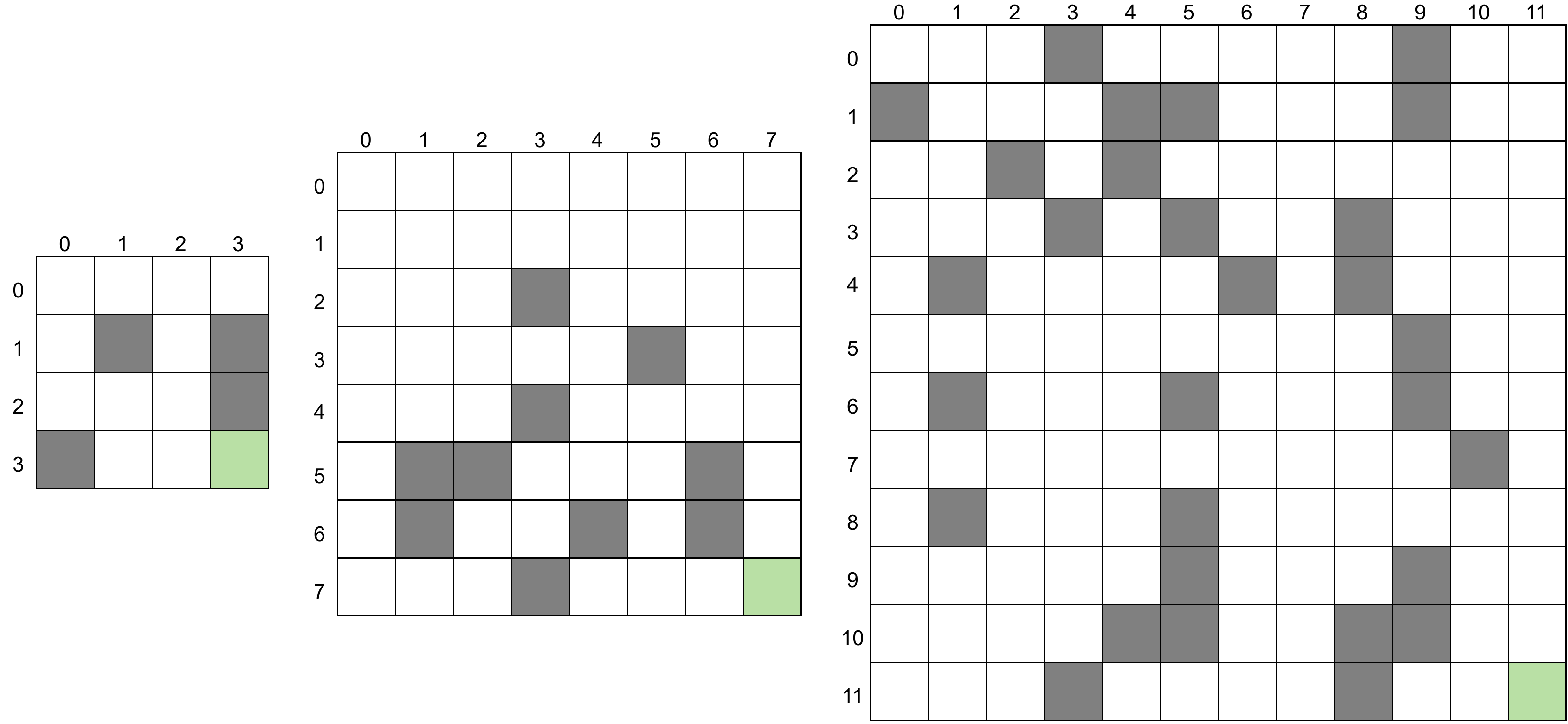}
    \caption{FrozenLake environments. Left to right: $\textit{M} = 4, 8, 12$. Cell legend: white = frozen, grey = hole, green = goal. States are $\bm{(\textit{x}, \textit{y})}$ coordinates: $\bm{\textit{x}}$ = column, $\bm{\textit{y}}$ = row.}
    \label{fig:fl_grids}
\end{figure}

We now apply our three learning systems across a suite of FrozenLake environments\footnote{Code used for our experiments is available at: \url{https://github.com/jtbish/gecco22}.} in order to make comments about how the different systems learn, and to test our Pittsburgh Strength Hypothesis.
Figure~\ref{fig:fl_grids} shows the structures of these environments, where we use $M$ to refer to the (square) grid size of each environment.
We refer the reader to \cite{bishop_optimality-based_2020} for a full description of FL.
An allocentric state representation is used because there are situations in each grid where using an egocentric representation would result in state aliasing.
This justifies our use of allocentric learning systems.
Each environment is parametrised by a value $\flsp$ that controls the amount of transition stochasticity: the probability of slipping in either direction perpendicular to the intended direction of movement.
In this work we consider the cases $\flsp \in \{0, 0.1, 0.3, 0.5\}$, i.e. one deterministic and three stochastic cases.
%Across three grid sizes and these $\flsp$ values we thus have a suite of twelve environments that become more difficult as $M$ and $\flsp$ increase.
%To summarise briefly: environment state is given as an $(x, y)$ tuple where both $x$ and $y$ inhabit the range $[0, M{-}1]$, $x$ is the column, and $y$ is the row of the grid starting in the top-left corner. 
%Continuing, $A = \{\text{Up}, \text{Down}, \text{Left}, \text{Right}\}$. A reward of +1 is given when the agent transitions into the goal state with 0 reward elsewhere, making the reward signal very sparse. Episodes end when the agent either i) falls into a hole, ii) reaches the goal, or iii) runs out of time.

This results in twelve environments, corresponding to each combination of $(M, \flsp)$, that become more difficult as $M$ and $\flsp$ increase.
We use this 2-tuple notation to refer to a specific environment.
Across all environments, we use a discount factor of $\gamma{=}0.95$.
$t_{\text{max}}$ values are set as $150, 300, 450$ for $M = 4, 8, 12$.
For each grid, we set its respective $S_I$ to be the set of all frozen cells, yielding $\vert S_I \vert$ values of $11, 53, 114$ for $M = 4, 8, 12$, and meaning we care about \textit{learning policies to navigate from every frozen cell to the goal}.
For the setup of XCS, as the environments are allocentric, we employ computed prediction (i.e. XCSF).
Specifically, we use linear computed prediction with the recursive least squares (RLS) prediction update algorithm \cite{lanzi_generalization_2007}. Further, as per the advice of Drugowitsch in \cite{drugowitsch_design_2008}, we incorporate a forgetting factor $\lambda_{\text{rls}}$ into RLS, due to the non-stationary nature of predicting the action-value function in an online fashion via temporal difference learning.
Parameters for RLS are $\delta_{\text{rls}}{=}10, \lambda_{\text{rls}}{=}0.999$.
Other (fixed) XCS parameters are: $\alpha{=}\beta{=}0.1, \epsilon_0{=}0.01, \nu{=}5, \theta_{\text{GA}}{=}\theta_{\text{del}}{=}\theta_{\text{sub}}{=}50, \tau{=}0.5$ (tournament selection \cite{butz2006rule} is used), $\chi{=}0.8, \mu{=}0.04, \delta{=}0.1, \epsilon_I{=}f_I{=}10^{-3}, x_0{=}10$.
In the case of stochastic environments, we incorporate the mechanism described by Lanzi in \cite{lanzi_extension_1999} for tracking the amount of predictive uncertainty induced by the environment.
This involves augmenting each classifier with a parameter $\mu$ (initialised to $\mu_I$) which is updated with a separate learning rate $\beta_\epsilon$, which we set as $\mu_I{=}10^{-3}, \beta_\epsilon{=}0.05$.
Depending on the grid size $M$, XCS uses differing population sizes $N$ and covering/mutation parameters.
We use the UBR rule representation for hyperrectangular conditions in XCS, as done for PPL (see Section~\ref{1sec:pitt_design_commonalities}).
Covering is controlled by $r_0$ and mutation amount is controlled by $m_0$ (see \cite{wilson_mining_2000}). These are set to $\frac{M}{2}$ and $\frac{M}{4}$ respectively.
Population sizes are set to $700, 2100, 4200$ for $M = 4, 8, 12$.

For PPL, we use common parameters for both variants of:\\ $\var{tournSize}{=}3, \var{pCross}{=}0.7, \var{pMut}{=}0.01$. 
$\var{idvSize}$ is set to $7, 21, 42$ for $M = 4, 8, 12$.
Population sizes are set to be sixteen times larger (ensuring even parity) than the corresponding $\var{idvSize}$: resulting in sizes of $112, 336, 672$.
PPL-ST additionally requires the specification of the $\var{numReinfRollouts}$ parameter. We performed two sensitivity analyses on this parameter to measure the final testing performance (FTP) of the system: the performance achieved after the training budget of $\var{numGens}$ generations was consumed.
For these analyses we considered $(8, 0)$ and $(8, 0.3)$ FL, with $\var{numGens}{=}250$ and $\var{numReinfRollouts}$ values of: $1, 2, 5, 10, 20, 50$.
Results (not reported) showed that this parameter had little impact on the overall FTP achieved by PPL-ST; even in the stochastic environment where it was expected the impact would be larger. However, it was observed that performance was more stable when selecting values $\ge 10$, after which no significant improvements were yielded.
Therefore, we use $\var{numReinfRollouts}{=}10$.
For PPL-ST we also set $\eta{=}0.1$, and like XCS, use $x_0{=}10$.

\begin{figure*}
    \centering
    \includegraphics[width=\textwidth]{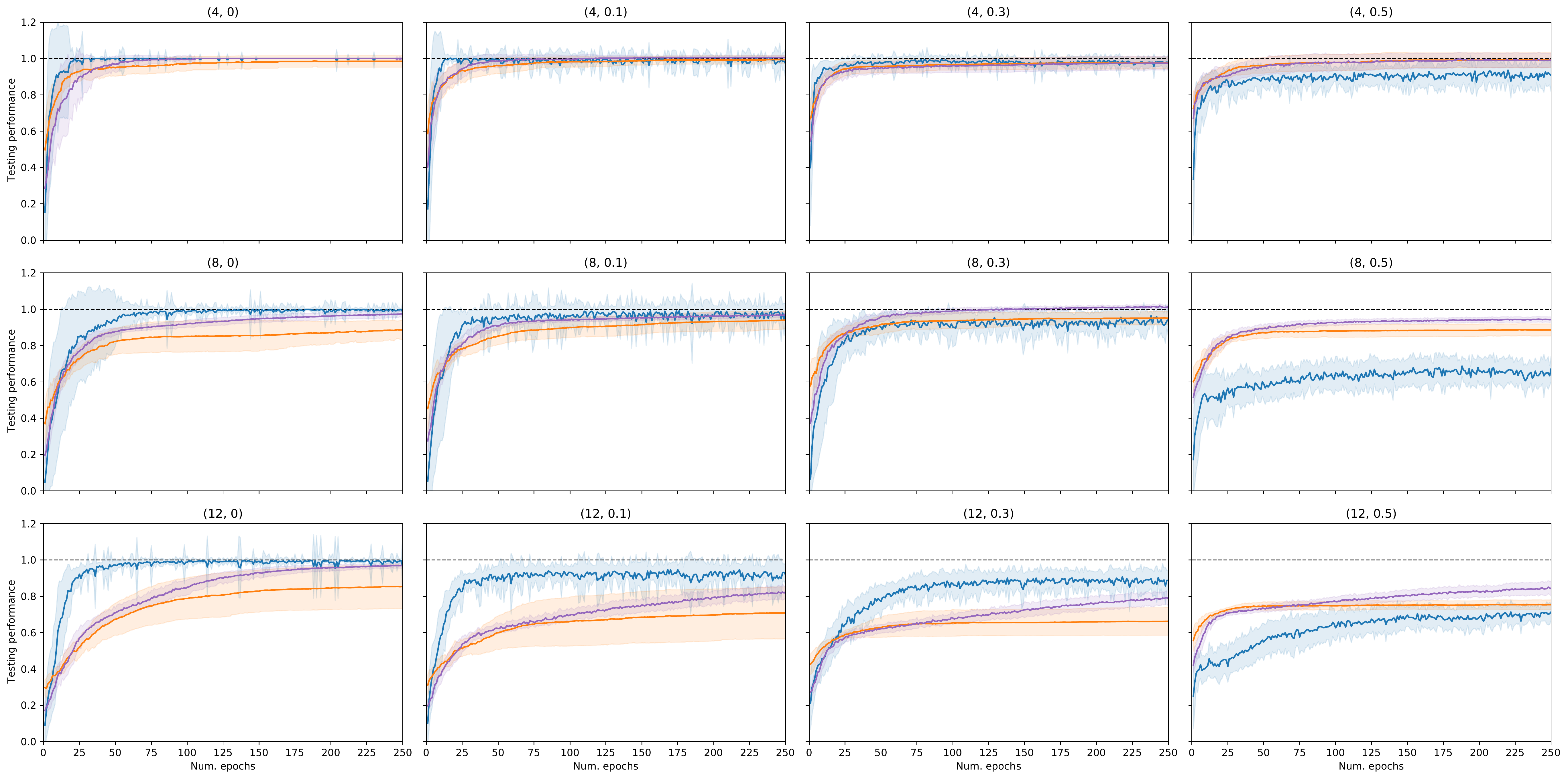}
    \caption{Performance achieved by XCS (blue), PPL-DL (orange), and PPL-ST (purple)  on FL environments with $\textit{M} \in \{4, 8, 12\}$ (rows, top to bottom), and $\textit{p}_{\text{slip}} \in \{0, 0.1, 0.3, 0.5\}$ (columns, left to right).
    Solid curves are the sample mean and shaded areas are the sample standard deviation, computed over thirty trials.
    All systems were run for 250 epochs on each environment.
    Vertical axes report the testing performance achieved by each system at each epoch, as a fraction of the optimal testing performance.} 
    \label{fig:perf_grid_plot}
\end{figure*}

A budget of $\var{numGens}{=}250$ is used for both PPL variants.
Because PPL population sizes are $112, 336, 672$ for $M = 4, 8, 12$, the respective epoch sizes are $\frac{112}{2}, \frac{336}{2}, \frac{672}{2} = 56, 168, 336$.
XCS is thus trained for 250 epochs, each epoch equivalent to a batch of $56, 168, 336$ GA invocations for $M = 4, 8, 12$.
When measuring system performance on deterministic environments, $z$ is set to be a single enumeration of $S_I$ (each initial state tested once). For stochastic environments it is necessary to test each initial state multiple times, so $z$ is assembled as $\mu$ enumerations of $S_I$ (we choose $\mu{=}30$); hence why $z$ is a sequence and not a set.
Performance curves for the three systems across all FL environments are shown in Figure~\ref{fig:perf_grid_plot}.
Thirty independent trials of each system were run.
In an effort to standardise interpretation of results over different environments, we report the testing performance of each system as a fraction of the optimal testing performance (OTP) for that environment; this is because OTP magnitude decreases as $M$ and $\flsp$ increase.
To calculate OTP, we performed value iteration on the MDP of each environment to produce an optimal policy, whose performance is equal to OTP.

\begin{figure}
    \centering
    \includegraphics[trim={0 0.33cm 0 0 },clip,width=\columnwidth]{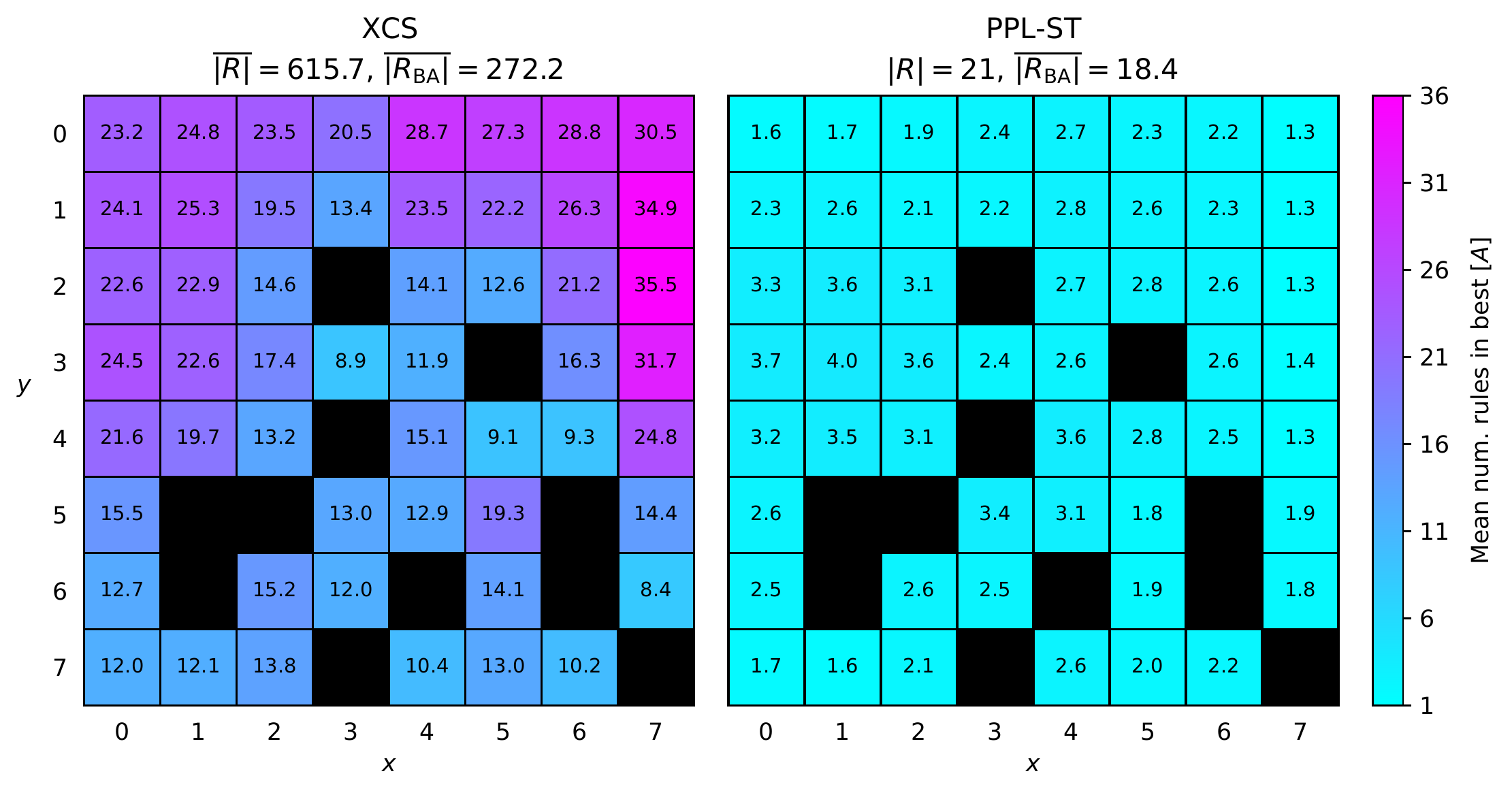}
    %\vspace{-0.5cm}
    \caption{Mean density of action sets that constitute the best action map for XCS and PPL-ST after 250 epochs in $(8, 0.3)$ FL.
    Terminal states (holes and goal) are coloured black.
    For each system, $\bm{\vert \textit{R} \vert}$ = size of the entire ruleset, and $\bm{\vert \textit{R}_{\text{BA}} \vert}$ = size of the ruleset that forms the best action map (overbars indicate sample means).
    For PPL-ST, $\bm{\vert \textit{R} \vert}$ is not a sample mean because fixed-length (constant size) rulesets are used.
    All statistics are calculated over thirty trials.}
    \vspace{-0.5cm}
    \label{fig:xcs_vs_pplst_action_set_heatmap}
\end{figure}

In Table~\ref{tab:etps_and_tests} we present the mean and standard deviation FTP achieved by each system in addition to the results of two Mann-Whitney U tests conducted for each environment, performed to determine if there were significant differences in FTP in the cases of: PPL-ST vs. XCS, and PPL-ST vs. PPL-DL. We do not consider the case of PPL-DL vs. XCS because PPL-DL is merely used as a baseline.
Mann-Whitney U tests were conducted instead of independent-sample $t$-tests because the vast majority of FTP samples failed a Shapiro-Wilk normality test ($\alpha{=}0.05$).
Lastly, in Figure~\ref{fig:xcs_vs_pplst_action_set_heatmap} we compare best action set densities and ruleset sizes of XCS and PPL-ST in $(8, 0.3)$ FL, in order to comment on the interpretability of the systems.
%This comparison shows, for each frozen cell, the mean number of rules that are part of best action sets (action sets that belong to the \textit{best action mapping} induced by the ruleset, i.e. its policy).

% Please add the following required packages to your document preamble:
% \usepackage{multirow}
% \usepackage[table,xcdraw]{xcolor}
% If you use beamer only pass "xcolor=table" option, i.e. \documentclass[xcolor=table]{beamer}
\begin{table}[]
\caption{$\bm{\textit{mean} \pm \textit{std}}$ final testing performance (achieved after 250 epochs) for PPL-ST (purple), PPL-DL (orange), and XCS (blue), across FL environments.
Two Mann-Whitney U tests were performed for each environment, comparing PPL-ST against PPL-DL and XCS, with family-wise $\alpha{=}0.05$ and Bonferroni correction.
Cells to the right-hand side of individual PPL-DL/XCS results describe our interpretation of its statistical comparison with PPL-ST, based on mean values shown.
If $\textit{H}_0$ was not rejected and no significant differences were found, $\bm{=}$ is shown, otherwise if $\textit{H}_0$ was rejected, a green $\bm{>}$/red $\bm{<}$ indicates PPL-ST performing \textit{better}/\textit{worse}, respectively.}
%\vspace{-0.3cm}
\label{tab:etps_and_tests}
\centering
\tiny
\setlength\tabcolsep{1.25pt}
% rowsep
\renewcommand{\arraystretch}{1}
\begin{tabular}{rclllllllllllll}
\multicolumn{1}{l}{} &
  \multicolumn{1}{l}{} &
  \multicolumn{10}{c}{$p_{\text{slip}}$} &
   &
   &
   \\
\multicolumn{1}{l}{} &
  \multicolumn{1}{l}{} &
  \multicolumn{1}{c}{0} &
   &
   &
  \multicolumn{1}{c}{0.1} &
   &
   &
  \multicolumn{1}{c}{0.3} &
   &
   &
  \multicolumn{1}{c}{0.5} &
   &
   &
   \\ \cline{3-3} \cline{6-6} \cline{9-9} \cline{12-12}
 &
  \multicolumn{1}{c|}{} &
  \multicolumn{1}{l|}{\cellcolor[HTML]{CCCDF7}$0.997 \pm 0.015$} &
   &
  \multicolumn{1}{l|}{} &
  \multicolumn{1}{l|}{\cellcolor[HTML]{CCCDF7}$1.006 \pm 0.008$} &
   &
  \multicolumn{1}{l|}{} &
  \multicolumn{1}{l|}{\cellcolor[HTML]{CCCDF7}$0.974 \pm 0.037$} &
   &
  \multicolumn{1}{l|}{} &
  \multicolumn{1}{l|}{\cellcolor[HTML]{CCCDF7}$0.991 \pm 0.039$} &
   &
   &
   \\ \cline{3-4} \cline{6-7} \cline{9-10} \cline{12-13}
 &
  \multicolumn{1}{c|}{} &
  \multicolumn{1}{l|}{\cellcolor[HTML]{F7D4B4}$0.985 \pm 0.032$} &
  \multicolumn{1}{l|}{{\color[HTML]{009901} \textbf{\textgreater{}}}} &
  \multicolumn{1}{l|}{} &
  \multicolumn{1}{l|}{\cellcolor[HTML]{F7D4B4}$0.994 \pm 0.021$} &
  \multicolumn{1}{l|}{{\color[HTML]{009901} \textbf{\textgreater{}}}} &
  \multicolumn{1}{l|}{} &
  \multicolumn{1}{l|}{\cellcolor[HTML]{F7D4B4}$0.977 \pm 0.033$} &
  \multicolumn{1}{l|}{\textbf{=}} &
  \multicolumn{1}{l|}{} &
  \multicolumn{1}{l|}{\cellcolor[HTML]{F7D4B4}$0.991 \pm 0.041$} &
  \multicolumn{1}{l|}{\textbf{=}} &
   &
   \\ \cline{3-4} \cline{6-7} \cline{9-10} \cline{12-13}
 &
  \multicolumn{1}{c|}{\multirow{-3}{*}{4}} &
  \multicolumn{1}{l|}{\cellcolor[HTML]{BBDCF3}$1.000 \pm 0.000$} &
  \multicolumn{1}{l|}{\textbf{=}} &
  \multicolumn{1}{l|}{} &
  \multicolumn{1}{l|}{\cellcolor[HTML]{BBDCF3}$0.973 \pm 0.083$} &
  \multicolumn{1}{l|}{{\color[HTML]{009901} \textbf{\textgreater{}}}} &
  \multicolumn{1}{l|}{} &
  \multicolumn{1}{l|}{\cellcolor[HTML]{BBDCF3}$0.979 \pm 0.046$} &
  \multicolumn{1}{l|}{\textbf{=}} &
  \multicolumn{1}{l|}{} &
  \multicolumn{1}{l|}{\cellcolor[HTML]{BBDCF3}$0.907 \pm 0.092$} &
  \multicolumn{1}{l|}{{\color[HTML]{009901} \textbf{\textgreater{}}}} &
   &
   \\ \cline{3-4} \cline{6-7} \cline{9-10} \cline{12-13}
 &
  \multicolumn{1}{l}{} &
   &
   &
   &
   &
   &
   &
   &
   &
   &
   &
   &
   &
   \\ \cline{3-3} \cline{6-6} \cline{9-9} \cline{12-12}
 &
  \multicolumn{1}{c|}{} &
  \multicolumn{1}{l|}{\cellcolor[HTML]{CCCDF7}$0.975 \pm 0.016$} &
   &
  \multicolumn{1}{l|}{} &
  \multicolumn{1}{l|}{\cellcolor[HTML]{CCCDF7}$0.968 \pm 0.012$} &
   &
  \multicolumn{1}{l|}{} &
  \multicolumn{1}{l|}{\cellcolor[HTML]{CCCDF7}$1.013 \pm 0.012$} &
   &
  \multicolumn{1}{l|}{} &
  \multicolumn{1}{l|}{\cellcolor[HTML]{CCCDF7}$0.943 \pm 0.015$} &
   &
   &
   \\ \cline{3-4} \cline{6-7} \cline{9-10} \cline{12-13}
 &
  \multicolumn{1}{c|}{} &
  \multicolumn{1}{l|}{\cellcolor[HTML]{F7D4B4}$0.887 \pm 0.053$} &
  \multicolumn{1}{l|}{{\color[HTML]{009901} \textbf{\textgreater{}}}} &
  \multicolumn{1}{l|}{} &
  \multicolumn{1}{l|}{\cellcolor[HTML]{F7D4B4}$0.941 \pm 0.048$} &
  \multicolumn{1}{l|}{{\color[HTML]{009901} \textbf{\textgreater{}}}} &
  \multicolumn{1}{l|}{} &
  \multicolumn{1}{l|}{\cellcolor[HTML]{F7D4B4}$0.951 \pm 0.037$} &
  \multicolumn{1}{l|}{{\color[HTML]{009901} \textbf{\textgreater{}}}} &
  \multicolumn{1}{l|}{} &
  \multicolumn{1}{l|}{\cellcolor[HTML]{F7D4B4}$0.886 \pm 0.033$} &
  \multicolumn{1}{l|}{{\color[HTML]{009901} \textbf{\textgreater{}}}} &
   &
   \\ \cline{3-4} \cline{6-7} \cline{9-10} \cline{12-13}
 &
  \multicolumn{1}{c|}{\multirow{-3}{*}{8}} &
  \multicolumn{1}{l|}{\cellcolor[HTML]{BBDCF3}$0.986 \pm 0.061$} &
  \multicolumn{1}{l|}{{\color[HTML]{FE0000} \textbf{\textless{}}}} &
  \multicolumn{1}{l|}{} &
  \multicolumn{1}{l|}{\cellcolor[HTML]{BBDCF3}$0.980 \pm 0.051$} &
  \multicolumn{1}{l|}{{\color[HTML]{FE0000} \textbf{\textless{}}}} &
  \multicolumn{1}{l|}{} &
  \multicolumn{1}{l|}{\cellcolor[HTML]{BBDCF3}$0.943 \pm 0.091$} &
  \multicolumn{1}{l|}{{\color[HTML]{009901} \textbf{\textgreater{}}}} &
  \multicolumn{1}{l|}{} &
  \multicolumn{1}{l|}{\cellcolor[HTML]{BBDCF3}$0.672 \pm 0.070$} &
  \multicolumn{1}{l|}{{\color[HTML]{009901} \textbf{\textgreater{}}}} &
   &
   \\ \cline{3-4} \cline{6-7} \cline{9-10} \cline{12-13}
 &
  \multicolumn{1}{r}{} &
   &
   &
   &
   &
   &
   &
   &
   &
   &
   &
   &
   &
   \\ \cline{3-3} \cline{6-6} \cline{9-9} \cline{12-12}
 &
  \multicolumn{1}{c|}{} &
  \multicolumn{1}{l|}{\cellcolor[HTML]{CCCDF7}$0.969 \pm 0.015$} &
   &
  \multicolumn{1}{l|}{} &
  \multicolumn{1}{l|}{\cellcolor[HTML]{CCCDF7}$0.825 \pm 0.034$} &
   &
  \multicolumn{1}{l|}{} &
  \multicolumn{1}{l|}{\cellcolor[HTML]{CCCDF7}$0.787 \pm 0.038$} &
   &
  \multicolumn{1}{l|}{} &
  \multicolumn{1}{l|}{\cellcolor[HTML]{CCCDF7}$0.842 \pm 0.041$} &
   &
   &
   \\ \cline{3-4} \cline{6-7} \cline{9-10} \cline{12-13}
 &
  \multicolumn{1}{c|}{} &
  \multicolumn{1}{l|}{\cellcolor[HTML]{F7D4B4}$0.854 \pm 0.121$} &
  \multicolumn{1}{l|}{{\color[HTML]{009901} \textbf{\textgreater{}}}} &
  \multicolumn{1}{l|}{} &
  \multicolumn{1}{l|}{\cellcolor[HTML]{F7D4B4}$0.709 \pm 0.143$} &
  \multicolumn{1}{l|}{{\color[HTML]{009901} \textbf{\textgreater{}}}} &
  \multicolumn{1}{l|}{} &
  \multicolumn{1}{l|}{\cellcolor[HTML]{F7D4B4}$0.663 \pm 0.077$} &
  \multicolumn{1}{l|}{{\color[HTML]{009901} \textbf{\textgreater{}}}} &
  \multicolumn{1}{l|}{} &
  \multicolumn{1}{l|}{\cellcolor[HTML]{F7D4B4}$0.754 \pm 0.027$} &
  \multicolumn{1}{l|}{{\color[HTML]{009901} \textbf{\textgreater{}}}} &
   &
   \\ \cline{3-4} \cline{6-7} \cline{9-10} \cline{12-13}
\multirow{-11}{*}{$M$} &
  \multicolumn{1}{c|}{\multirow{-3}{*}{12}} &
  \multicolumn{1}{l|}{\cellcolor[HTML]{BBDCF3}$0.962 \pm 0.176$} &
  \multicolumn{1}{l|}{{\color[HTML]{009901} \textbf{\textgreater{}}}} &
  \multicolumn{1}{l|}{} &
  \multicolumn{1}{l|}{\cellcolor[HTML]{BBDCF3}$0.920 \pm 0.055$} &
  \multicolumn{1}{l|}{{\color[HTML]{FE0000} \textbf{\textless{}}}} &
  \multicolumn{1}{l|}{} &
  \multicolumn{1}{l|}{\cellcolor[HTML]{BBDCF3}$0.888 \pm 0.053$} &
  \multicolumn{1}{l|}{{\color[HTML]{FE0000} \textbf{\textless{}}}} &
  \multicolumn{1}{l|}{} &
  \multicolumn{1}{l|}{\cellcolor[HTML]{BBDCF3}$0.714 \pm 0.046$} &
  \multicolumn{1}{l|}{{\color[HTML]{009901} \textbf{\textgreater{}}}} &
   &
   \\ \cline{3-4} \cline{6-7} \cline{9-10} \cline{12-13}
\end{tabular}
\vspace{-0.3cm}
\end{table}

%trim={<left> <lower> <right> <upper>}

\section{Discussion}
\label{sec:discussion}
An initial point of discussion is that the $(4, 0.1)$ and $(8, 0.3)$ environments show mean performance values $> 1$ for PPL-ST, see Figure~\ref{fig:perf_grid_plot} and Table~\ref{tab:etps_and_tests}, even though we have stipulated that performance values are a fraction of OTP, which is calculated for stochastic environments over a \textit{finite number of initial states}, and thus uses a \textit{sample of possible transitions that could occur in the environment}.
It is possible that our learning systems can ``overfit'' to the empirical transition noise present in the performance evaluation, thus outperforming OTP.
This is not a concern as it reflects high performance and a quirk of sampling in an MDP.

Focusing on XCS in Figure~\ref{fig:perf_grid_plot}, we see that as the environments become more difficult (moving from top-left to bottom right), the FTP of XCS generally decreases.
Thus we can infer that XCS is unable to cope very well with increased action chain lengths in tandem with increased transition stochasticity.
Further, in stochastic environments XCS exhibits a non-trivial amount of variation in its performance as it does not converge to a static value.
In contrast to XCS, the mean performance of PPL-DL improves in a smooth fashion in all environments.
In certain environments ($(8, 0), (12, 0), (12, 0.1),$ and $(12, 0.3)$) PPL-DL also exhibits quite a large amount of performance variation, sometimes more than XCS.
Interestingly, the variation shrinks when $\flsp{=}0.5$ across all grid sizes.
Like PPL-DL, the mean performance of PPL-ST also increases smoothly as training progresses, and additionally PPL-ST exhibits the smallest amount of performance variation of the three systems, which is a desirable trait.
Particularly in the $M{=}12$ environments, PPL-ST does not appear to have converged, suggesting that it \textit{ideally} should have been run for longer.

This returns us to the assertion made that the computational effort required for running Pittsburgh systems is less problematic given modern computing power.
However, the computational complexity of our new Pittsburgh systems remains high.
Considering our use of PPL-ST, we set $\var{numReinfRollouts}{=}10$, and for a stochastic environment with $M{=}12$, we have $\var{popSize}{=}672$, and $\vert z \vert = 114 \cdot 30 = 3420$.
Applying Equation~\ref{eqn:pplst_epsPerGen}, in such a case the system consumed $672 \cdot (10 + 3420) = 2{,}304{,}960$ episodes per generation. Thus, over the entire 250 generations, $250 \cdot 2{,}304{,}960 = 576{,}240{,}000$ (roughly half a billion) episodes were required.
This figure is multiple orders of magnitude larger than the number of episodes used by XCS for an equivalent 250 epochs.
A redeeming feature of PPL is that it is trivially parallelisable over multiple processors.
Further methods could possibly be incorporated into PPL to reduce its computational complexity; one being the stratification process used by GAssist and BioHEL, where fitness computation only uses a particular stratum of the training data at each generation.

We now quantitatively address the PSH (see Section~\ref{sec:intro}) using the results in Table~\ref{tab:etps_and_tests}.
PPL-ST performs either equivalently ($M{=}4$) or better than PPL-DL across all domains.
Thus, as $M$ and $\flsp$ increase, we can say that PPL-ST \textit{does} consistently outperform PPL-DL, and so we accept part i) of the PSH.
This suggests that the additional strength-based mechanisms of PPL-ST result in increased performance compared to its zeroth-level counterpart.
The hypothesis test results for PPL-ST vs. XCS are more mixed.
Half of the time ($\frac{6}{12}$ environments), PPL-ST outperforms XCS, in two instances they are considered the same, and in the remaining four cases PPL-ST is outperformed by XCS.
Overall PPL-ST \textit{does not} consistently outperform XCS as both $M$ and $\flsp$ increase (notably $(12, 0.1)$ and $(12, 0.3)$), and therefore we must reject part ii) of the PSH.
However, we highlight that PPL-ST outperforms XCS on all grid sizes when $\flsp{=}0.5$.
This leads to an inference of PPL-ST being more adept than XCS at handling high levels of environmental uncertainty.

Finally, the interpretability of XCS and PPL-ST rulesets is discussed based on ``best action set'' density, see Figure~\ref{fig:xcs_vs_pplst_action_set_heatmap}, instead of ``match set'' density.
This is because XCS would necessarily have higher match set densities than PPL-ST because it constructs a complete map rather than a partial map over $S \times A$.
The density of best action sets gives an indication of the amount of overlap/redundancy present in encoding the policy.
XCS has much higher best action set density than PPL-ST across all states in the environment, indicating a high degree of overlap.
Further, the mean size of the ruleset constituting the best action map is much higher for XCS than for PPL-ST ($272.2$ vs. $18.4$ respectively).
In addition, note that in this particular environment PPL-ST outperforms XCS (see Table~\ref{tab:etps_and_tests}).

Similar best action set analyses for all other environments are included in the supplementary material.
Overall, these results indicate that the best action maps produced by PPL-ST are much more parsimonious than those of XCS, enabling interpretability.
One limitation of these results is that the chosen hyperrectangular rule encoding (UBR) scales poorly with the state dimensionality $d$, requiring $d$ intervals to be specified.
Thus, for large $d$, even though PPL-ST can produce parsimonious rulesets, the interpretability of individual rules may suffer.

\section{Conclusion}
\label{sec:conclusion}
% JBISH original:
% We presented two novel Pittsburgh LCSs for RL domains: PPL-DL and PPL-ST.
%The former was employed as a zeroth-level baseline system, and the latter was constructed as a simplified version of the Pittsburgh LCS SAMUEL, revisiting the use of Monte Carlo reinforcement learning methods for estimating rule utility.
%A literature review revealed that work comparing Pittsburgh and Michigan systems in RL domains is lacking, so we compared our two systems against the Michigan system XCS on a challenging suite of FrozenLake environments.
%Our findings show that PPL-ST is able to perform at or above the level of PPL-DL in all environments, which gives credence to the usefulness of its Monte Carlo learning techniques.
%PPL-ST did not consistently outperform XCS across all environments, but our findings do suggest that PPL-ST is the system most suited to problems that contain high levels of environmental uncertainty, and that XCS struggles to achieve high performance in problems that combine long action chains and environmental uncertainty.
%Additionally, we exemplified how PPL-ST is able to produce much more parsimonious (and therefore interpretable) rulesets than XCS while still achieving the same level of performance, which makes it a valuable algorithm for eXplainable AI in RL domains.
The main contributions are two novel Pittsburgh LCSs for RL domains, PPL-DL and PPL-ST, in addition to an experimental framework to fairly compare Michigan and Pittsburgh systems in RL domains.
PPL-DL was employed as a zeroth-level baseline system, while PPL-ST was constructed as a simplified version of the Pittsburgh LCS SAMUEL, revisiting the use of MC RL methods for estimating rule strength. 
The previously limited work comparing Pittsburgh and Michigan systems in RL domains has been enhanced by comparing our two PPL variants against the Michigan system XCS on a challenging suite of FrozenLake environments.
Our findings show that PPL-ST can perform at or above the level of PPL-DL in all environments, which gives credence to the effectiveness of its learning techniques.
PPL-ST did not consistently outperform XCS across all environments, but our findings indicate that PPL-ST is the system most suited to problems that contain high levels of environmental uncertainty, and that XCS struggles to achieve high performance in problems that combine long action chains and environmental uncertainty. 
Lastly, we exemplified how PPL-ST can produce much more parsimonious (and therefore interpretable) rulesets than XCS while in some cases achieving better performance, making it a valuable algorithm for XAI in RL domains.
A promising direction for future work is revisiting the Lamarckian aspects of SAMUEL to further enhance PPL-ST, as well as considering how best to convert PPL-ST to use variable-length rulesets.

%%
%% The acknowledgments section is defined using the "acks" environment
%% (and NOT an unnumbered section). This ensures the proper
%% identification of the section in the article metadata, and the
%% consistent spelling of the heading.

%%
%% The next two lines define the bibliography style to be used, and
%% the bibliography file.
\bibliography{gecco2022.bib}
\bibliographystyle{ieeetr}

%%
%% If your work has an appendix, this is the place to put it.
\appendix

\newpage
\onecolumn
% Reset figure counters and give S prefix
\setcounter{figure}{0}
\setcounter{table}{0}
\makeatletter 
\renewcommand{\thefigure}{A\@arabic\c@figure}
\renewcommand{\thetable}{A\@arabic\c@table}
\makeatother

\noindent\begin{huge}\textbf{Supplementary Material}\end{huge}\\

\section{Additional Results}

\begin{figure}
\centering
\begin{subfigure}{0.425\textwidth}
    \includegraphics[width=\textwidth]{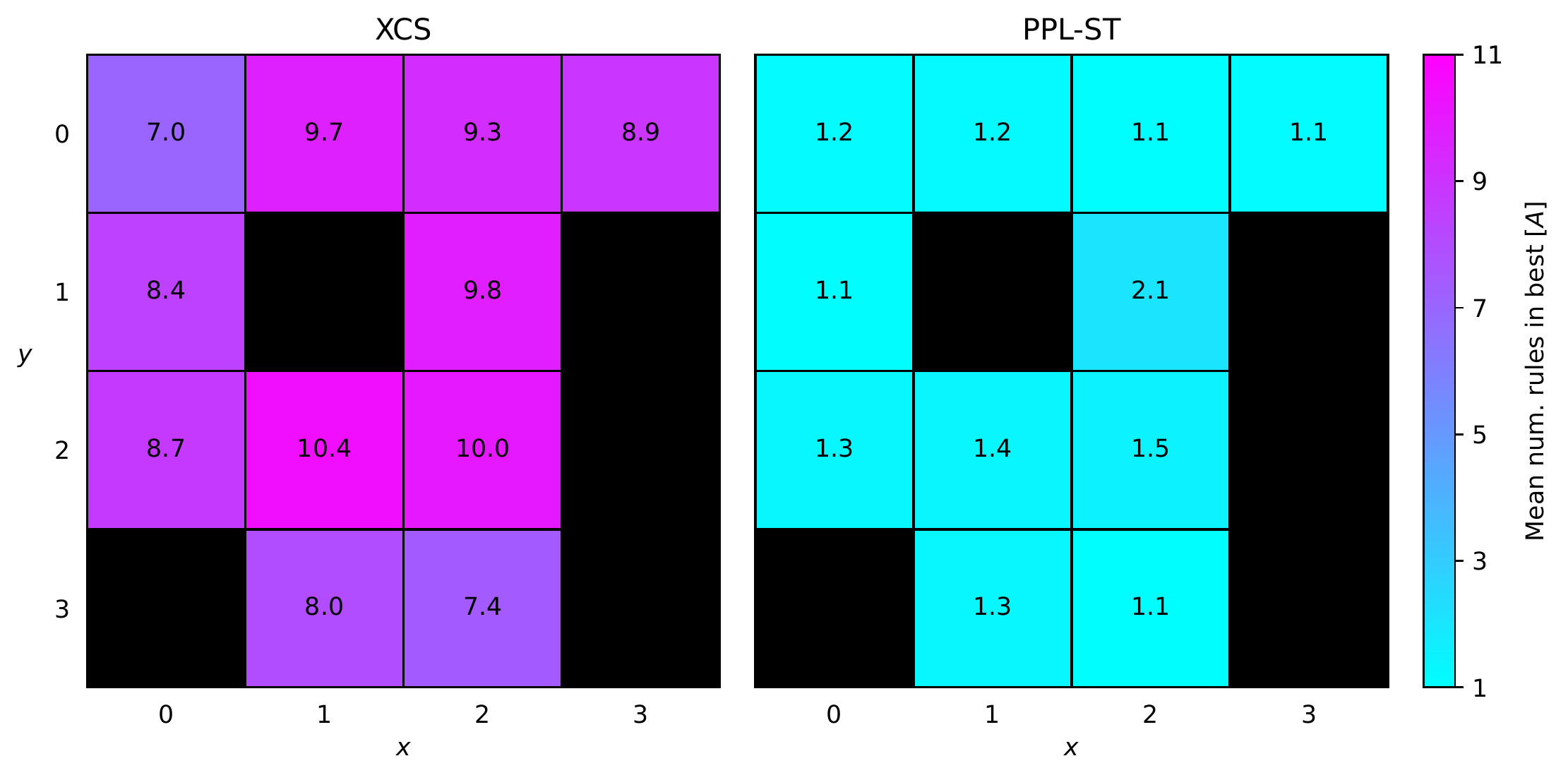}
    \caption{Local scale}
    \label{2fig:xcs_vs_pplst_bam_4_0_local}
\end{subfigure}
\hfill
\begin{subfigure}{0.425\textwidth}
    \includegraphics[width=\textwidth]{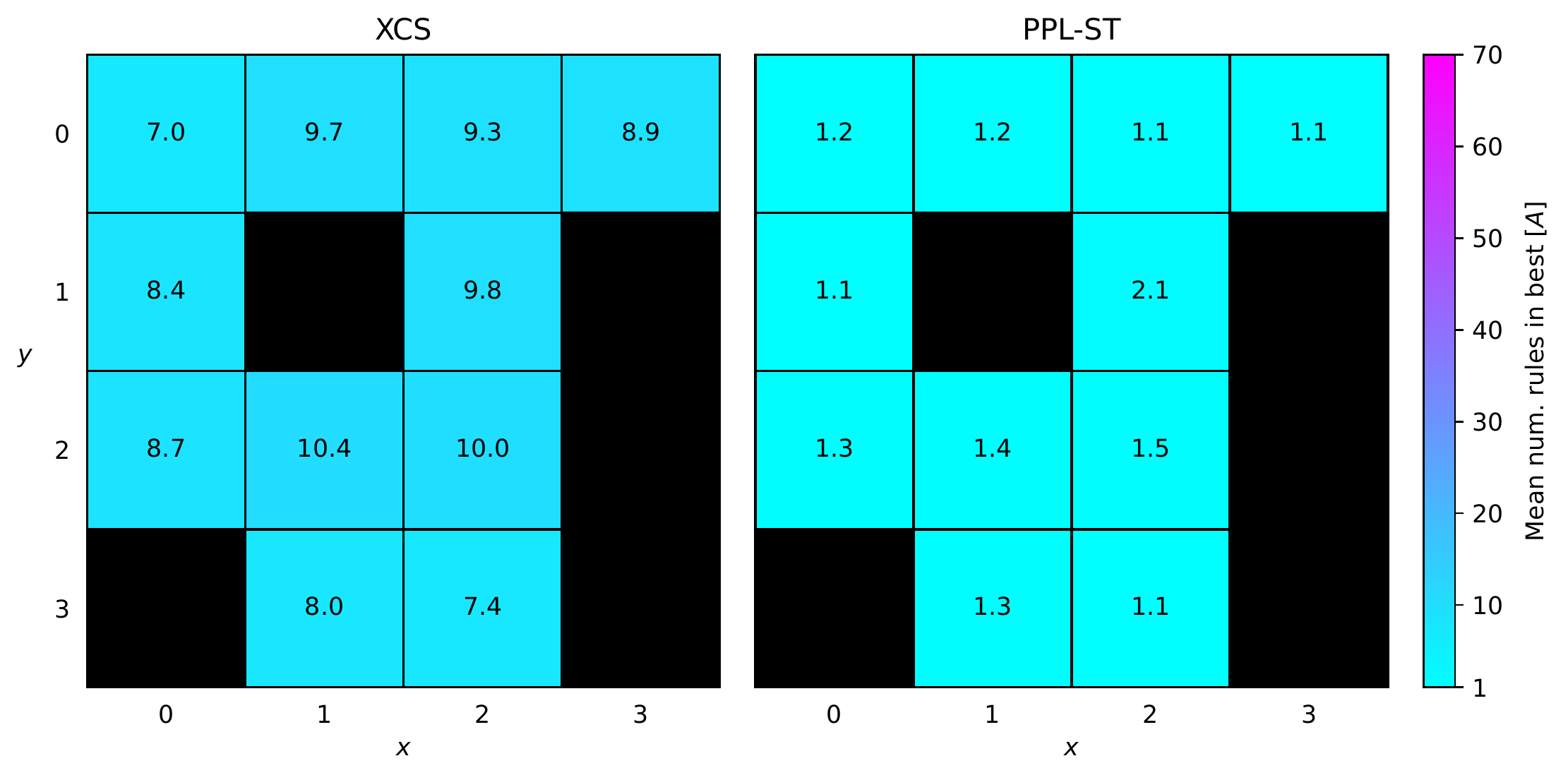}
    \caption{Global scale}
    \label{2fig:xcs_vs_pplst_bam_4_0_global}
\end{subfigure}
\caption{Mean best action set density of XCS and PPL-ST in $(4, 0)$ FL after 250 epochs.}
\label{fig:xcs_vs_pplst_bam_4_0_duo}
\end{figure}

\begin{figure}
\centering
\begin{subfigure}{0.425\textwidth}
    \includegraphics[width=\textwidth]{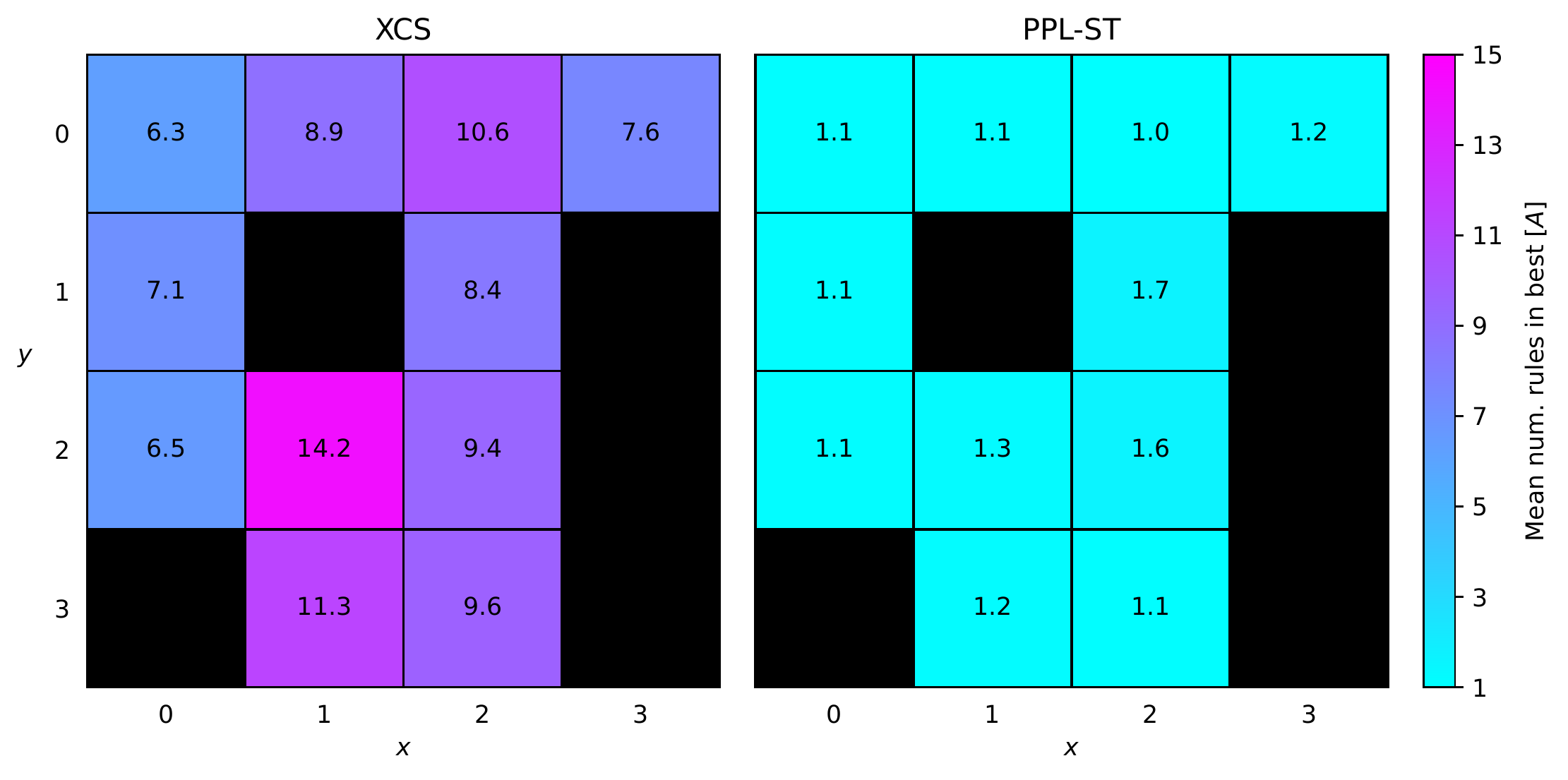}
    \caption{Local scale}
    \label{2fig:xcs_vs_pplst_bam_4_0.1_local}
\end{subfigure}
\hfill
\begin{subfigure}{0.425\textwidth}
    \includegraphics[width=\textwidth]{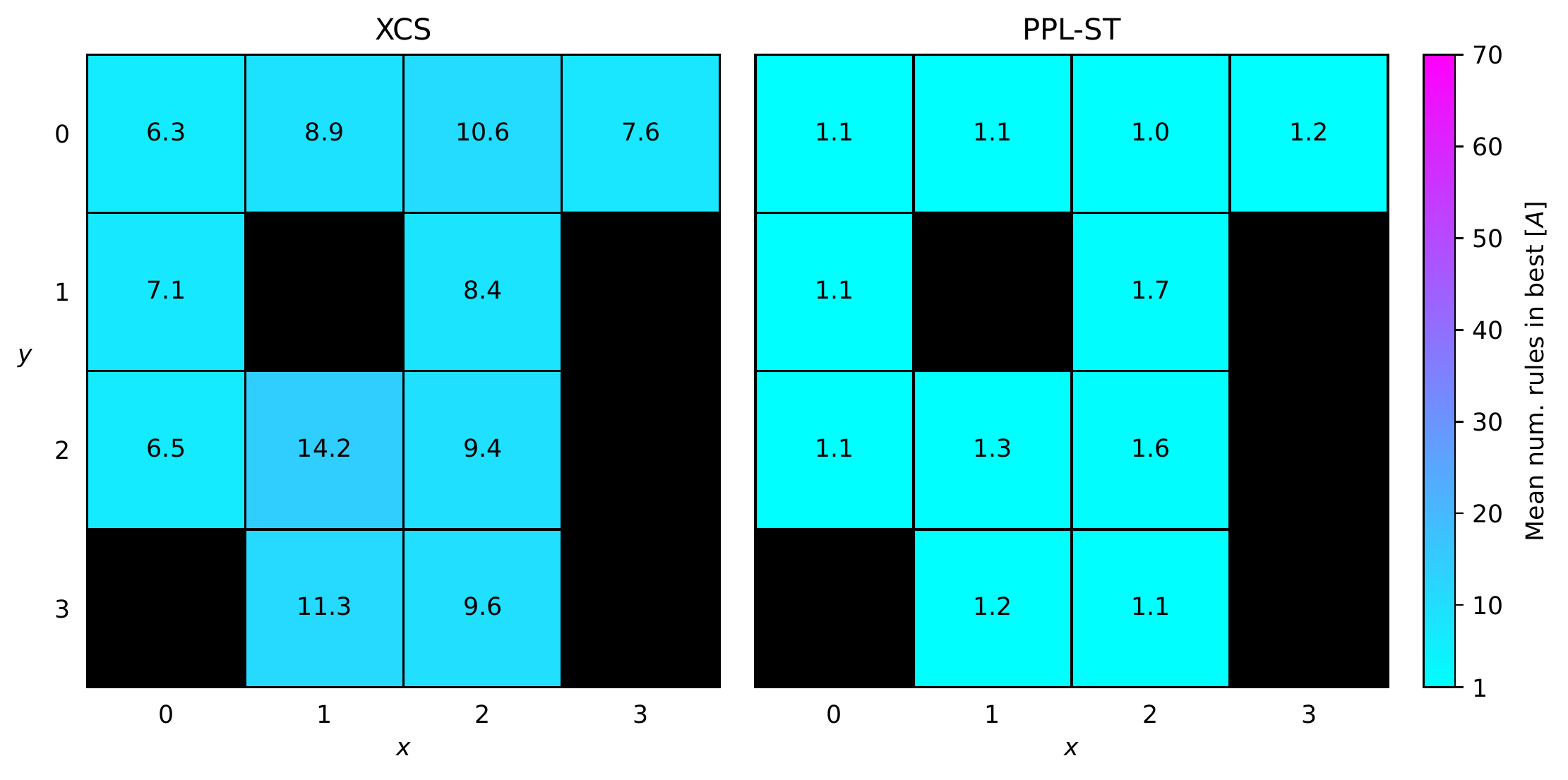}
    \caption{Global scale}
    \label{2fig:xcs_vs_pplst_bam_4_0.1_global}
\end{subfigure}
\caption{Mean best action set density of XCS and PPL-ST in $(4, 0.1)$ FL after 250 epochs.}
\label{fig:xcs_vs_pplst_bam_4_0.1_duo}
\end{figure}

\begin{figure}
\centering
\begin{subfigure}{0.425\textwidth}
    \includegraphics[width=\textwidth]{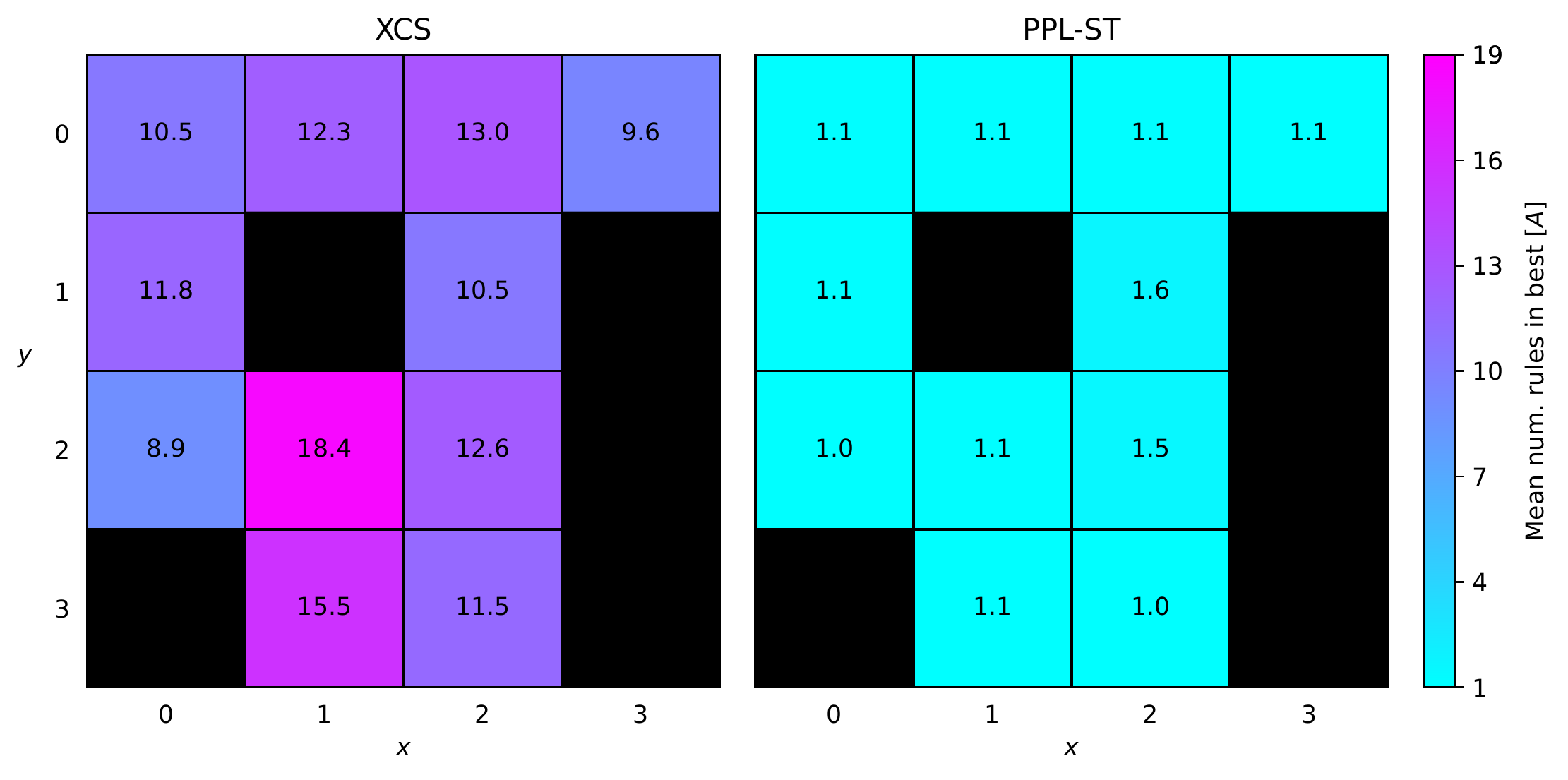}
    \caption{Local scale}
    \label{2fig:xcs_vs_pplst_bam_4_0.3_local}
\end{subfigure}
\hfill
\begin{subfigure}{0.425\textwidth}
    \includegraphics[width=\textwidth]{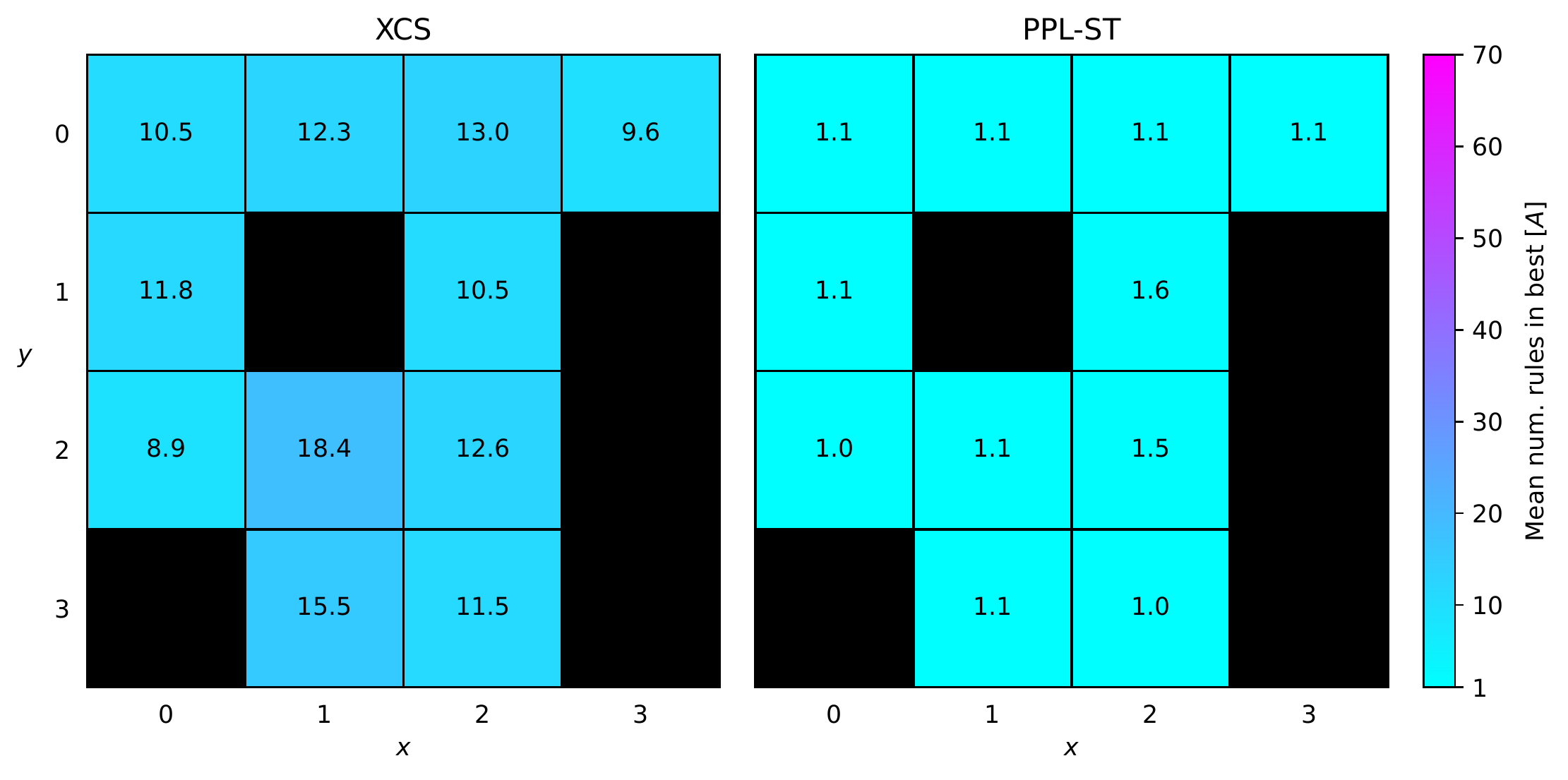}
    \caption{Global scale}
    \label{2fig:xcs_vs_pplst_bam_4_0.3_global}
\end{subfigure}
\caption{Mean best action set density of XCS and PPL-ST in $(4, 0.3)$ FL after 250 epochs.}
\label{fig:xcs_vs_pplst_bam_4_0.3_duo}
\end{figure}

\begin{figure}
\centering
\begin{subfigure}{0.425\textwidth}
    \includegraphics[width=\textwidth]{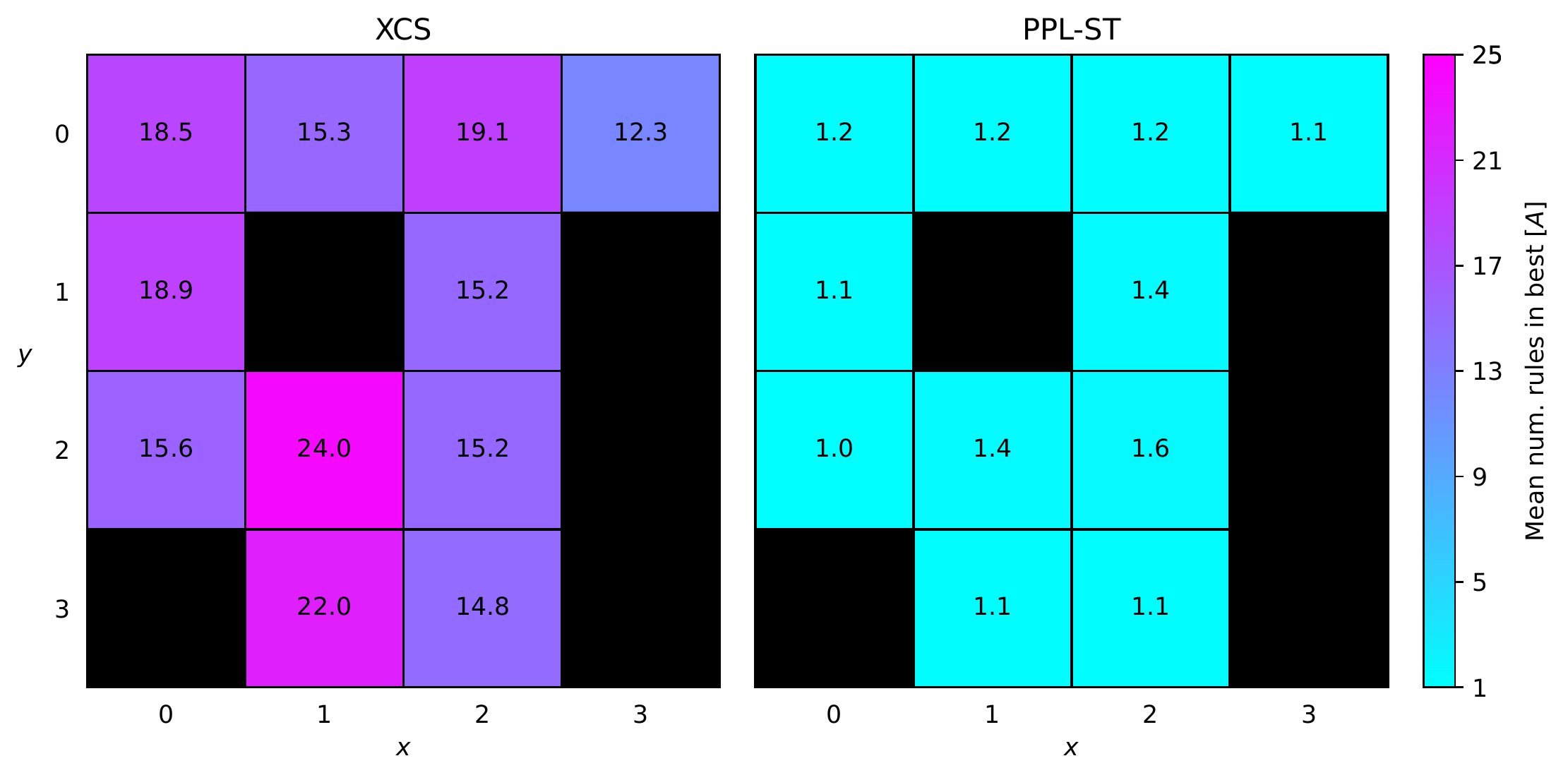}
    \caption{Local scale}
    \label{2fig:xcs_vs_pplst_bam_4_0.5_local}
\end{subfigure}
\hfill
\begin{subfigure}{0.425\textwidth}
    \includegraphics[width=\textwidth]{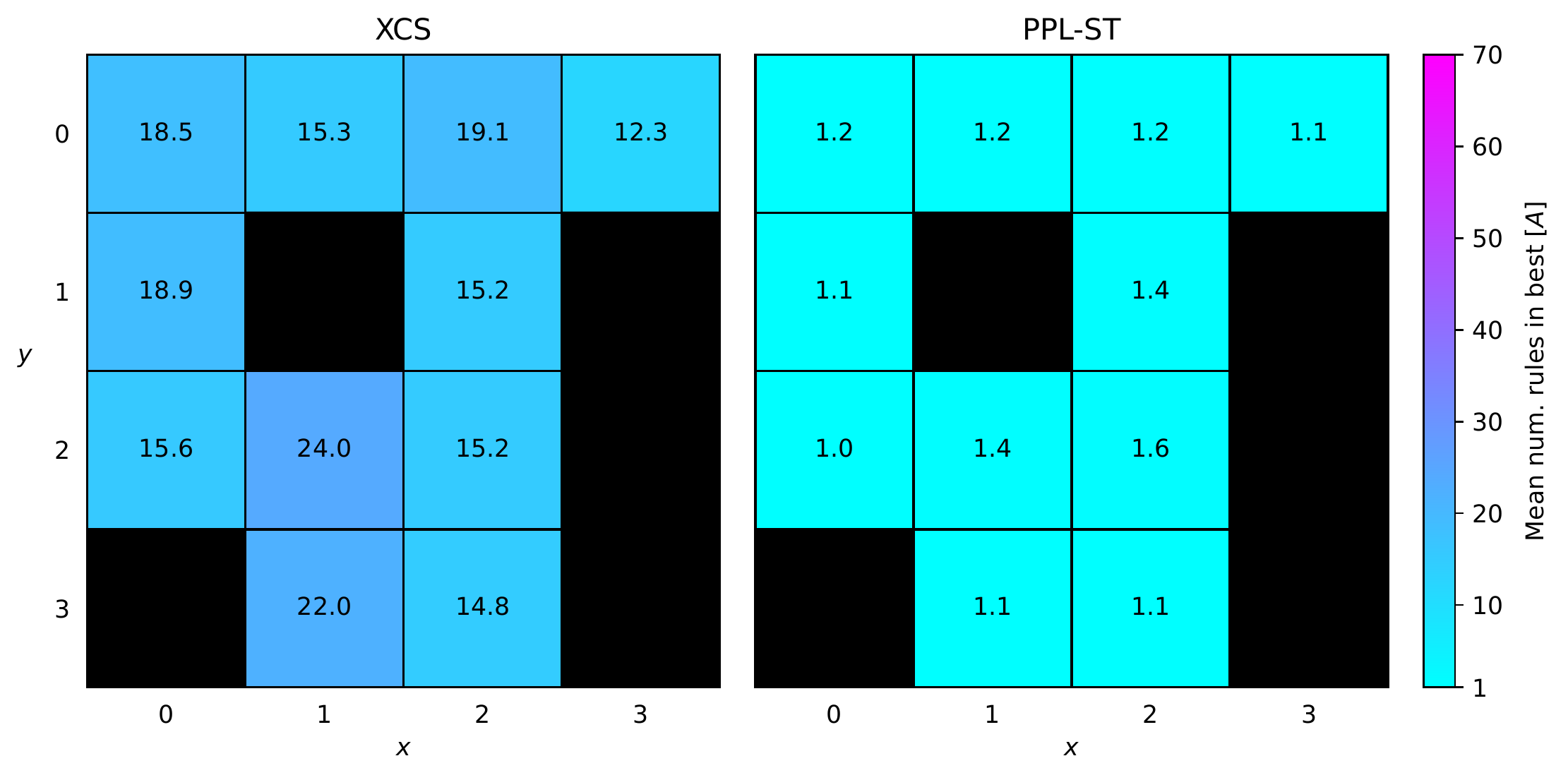}
    \caption{Global scale}
    \label{2fig:xcs_vs_pplst_bam_4_0.5_global}
\end{subfigure}
\caption{Mean best action set density of XCS and PPL-ST in $(4, 0.5)$ FL after 250 epochs.}
\label{fig:xcs_vs_pplst_bam_4_0.5_duo}
\end{figure}

\begin{figure}
\centering
\begin{subfigure}{0.425\textwidth}
    \includegraphics[width=\textwidth]{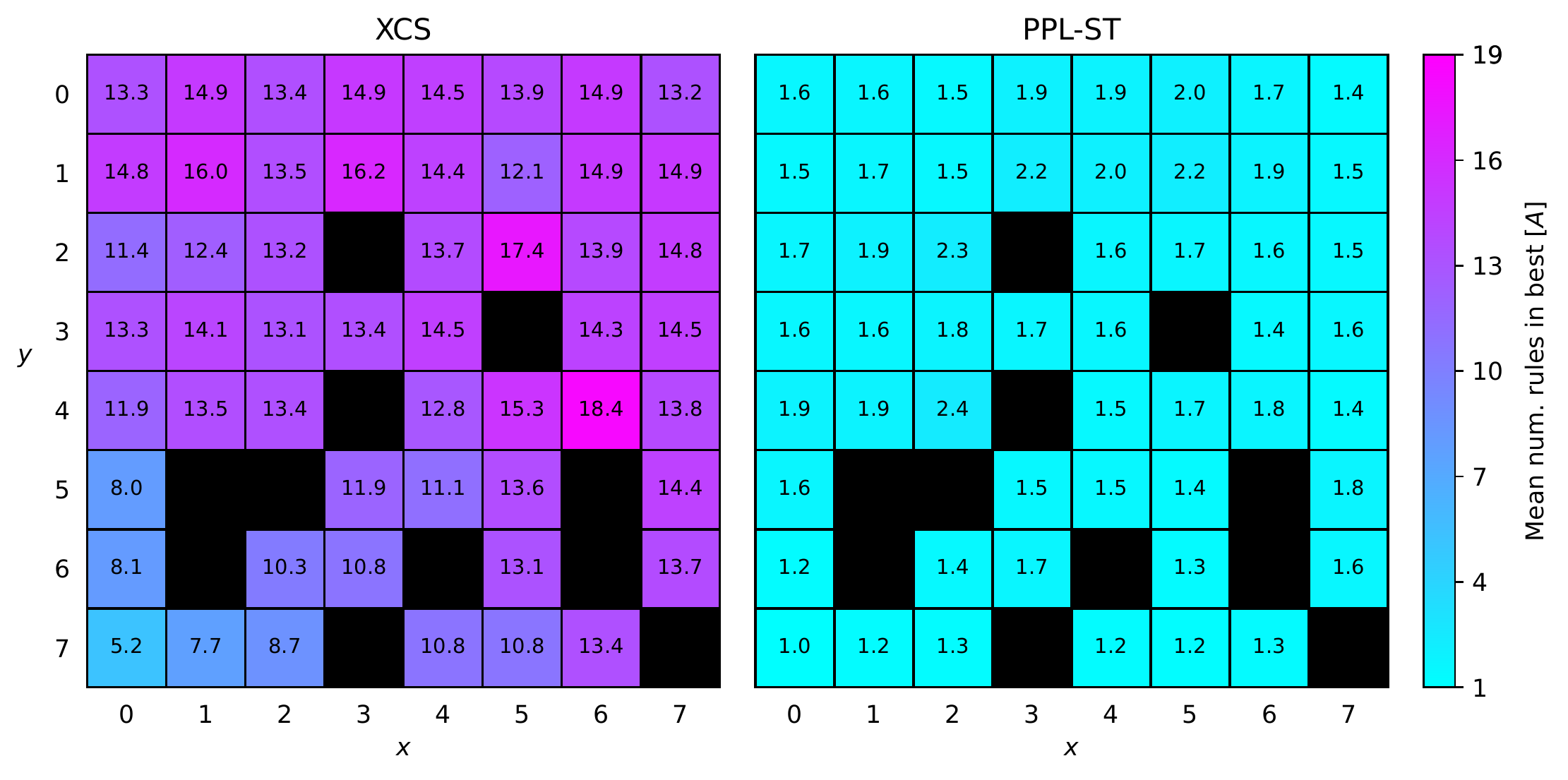}
    \caption{Local scale}
    \label{2fig:xcs_vs_pplst_bam_8_0_local}
\end{subfigure}
\hfill
\begin{subfigure}{0.425\textwidth}
    \includegraphics[width=\textwidth]{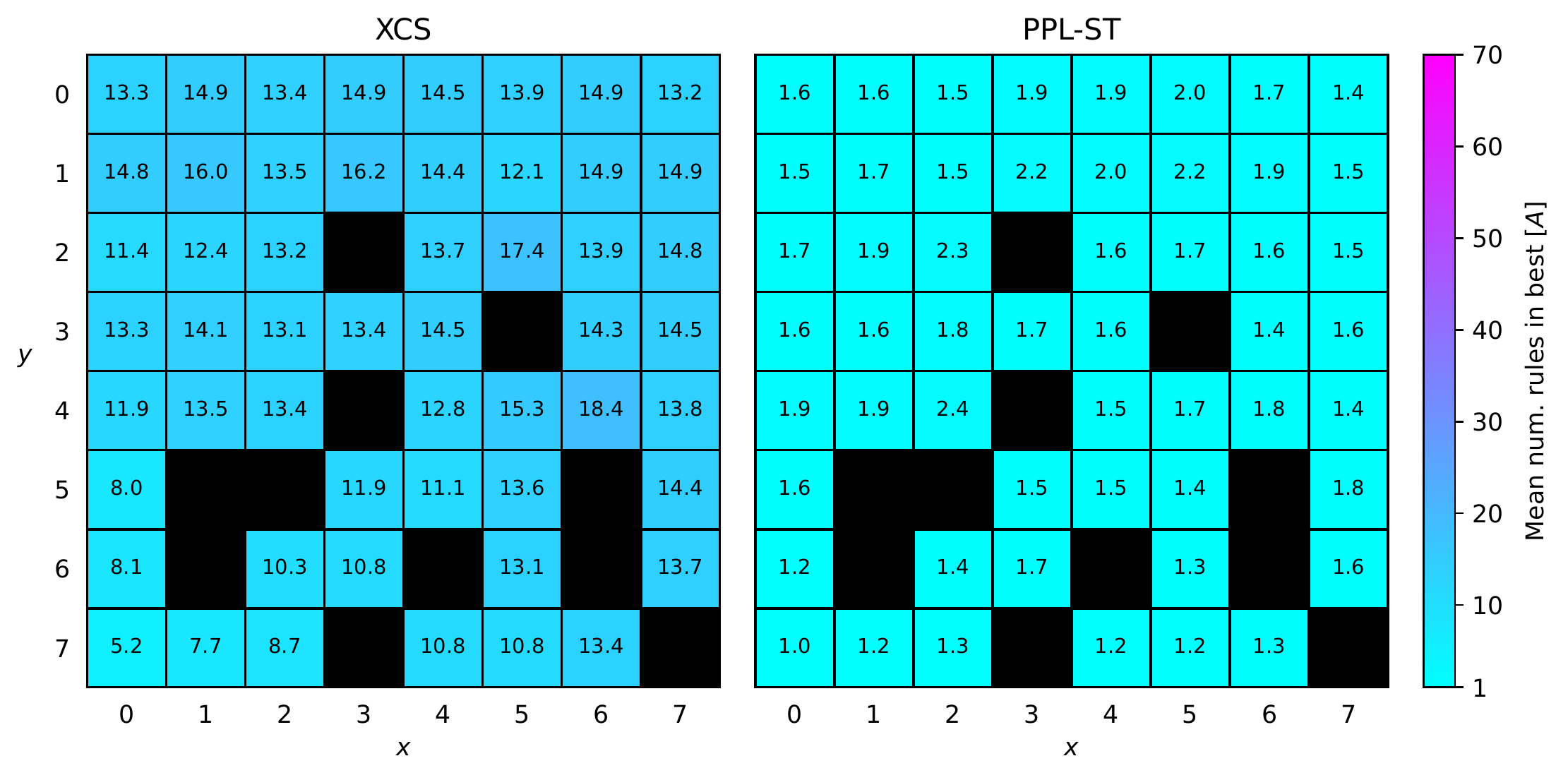}
    \caption{Global scale}
    \label{2fig:xcs_vs_pplst_bam_8_0_global}
\end{subfigure}
\caption{Mean best action set density of XCS and PPL-ST in $(8, 0)$ FL after 250 epochs.}
\label{fig:xcs_vs_pplst_bam_8_0_duo}
\end{figure}

\begin{figure}
\centering
\begin{subfigure}{0.425\textwidth}
    \includegraphics[width=\textwidth]{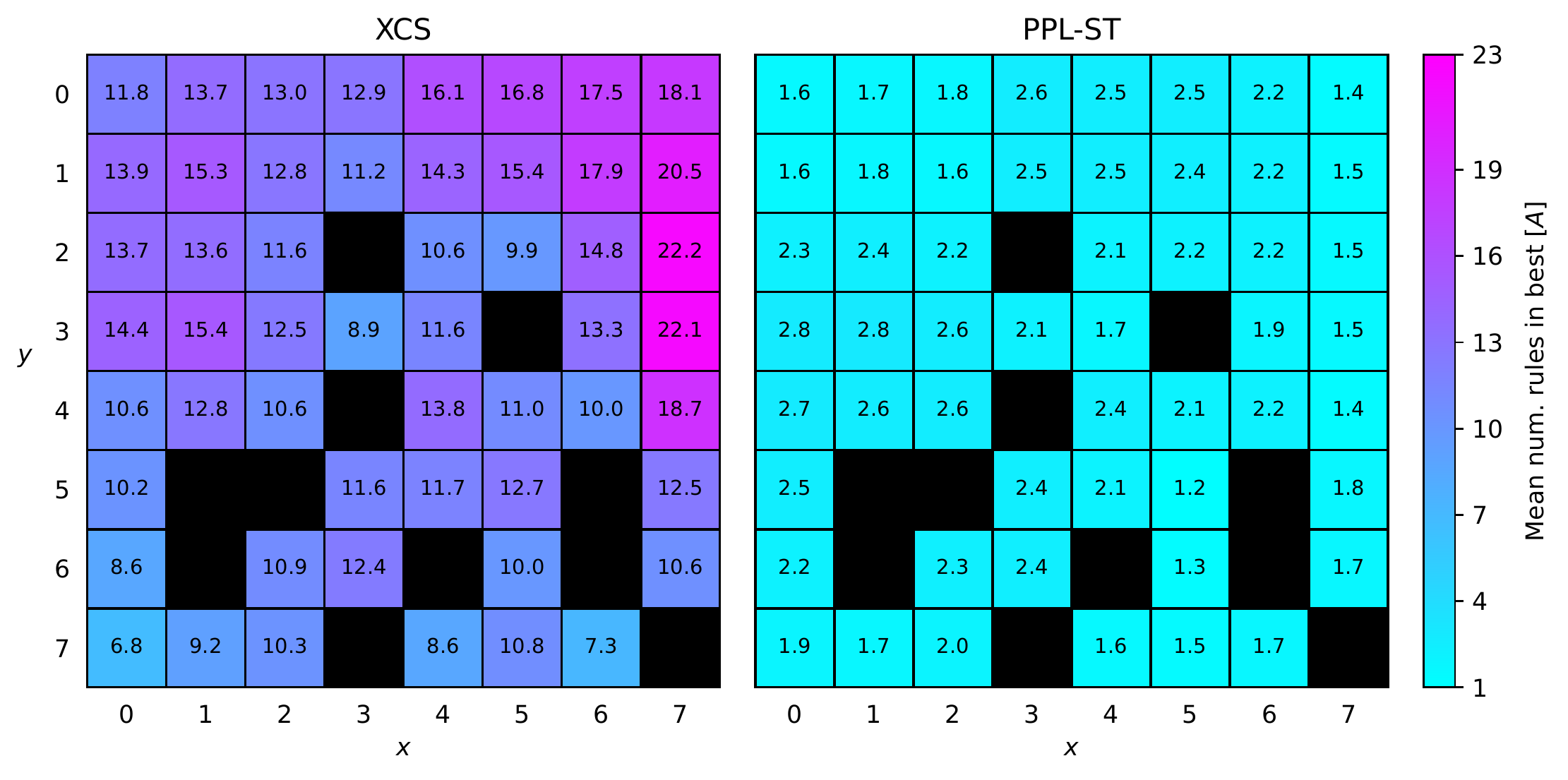}
    \caption{Local scale}
    \label{2fig:xcs_vs_pplst_bam_8_0.1_local}
\end{subfigure}
\hfill
\begin{subfigure}{0.425\textwidth}
    \includegraphics[width=\textwidth]{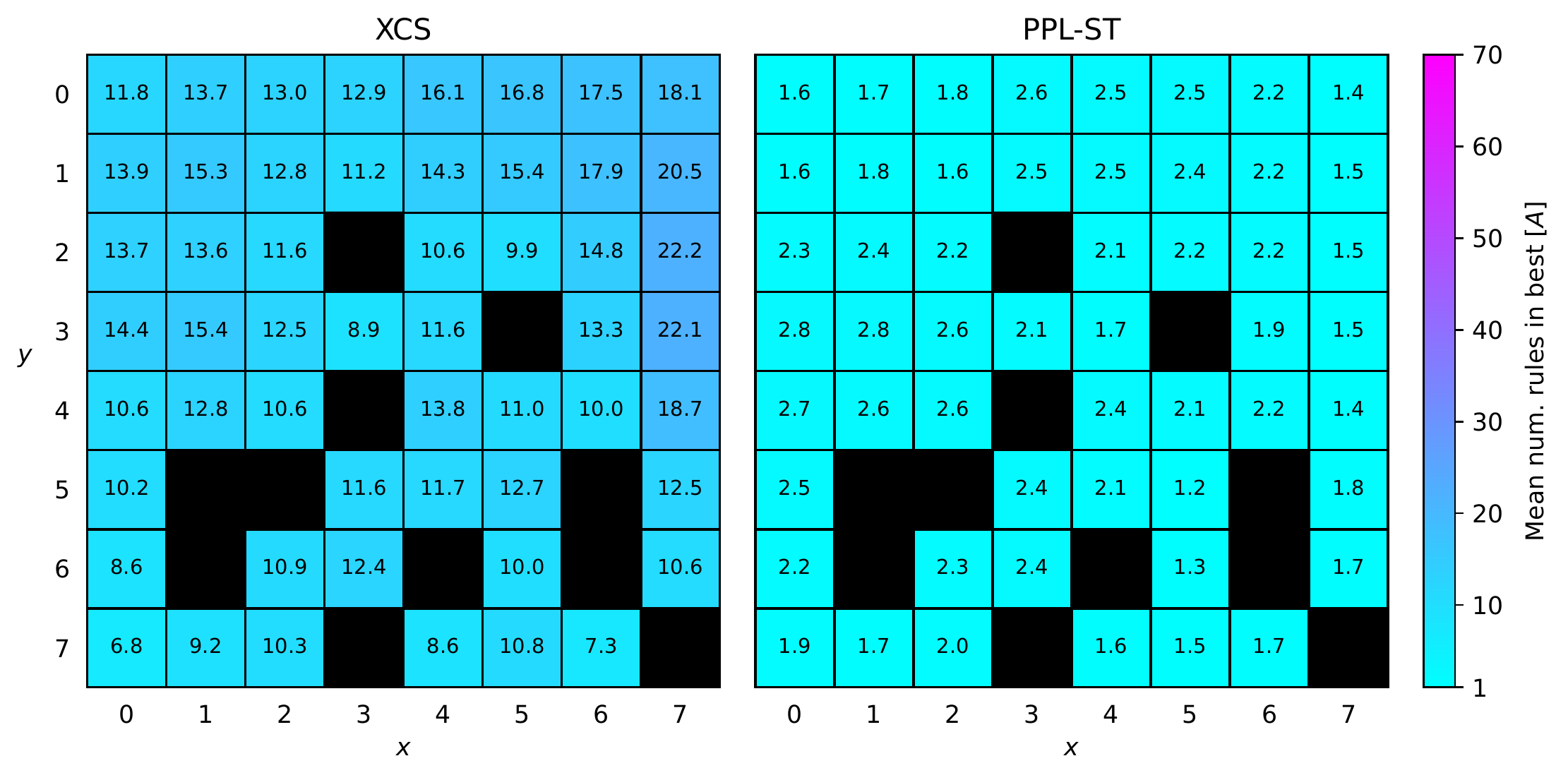}
    \caption{Global scale}
    \label{2fig:xcs_vs_pplst_bam_8_0.1_global}
\end{subfigure}
\caption{Mean best action set density of XCS and PPL-ST in $(8, 0.1)$ FL after 250 epochs.}
\label{fig:xcs_vs_pplst_bam_8_0.1_duo}
\end{figure}

\begin{figure}
\centering
\begin{subfigure}{0.425\textwidth}
    \includegraphics[width=\textwidth]{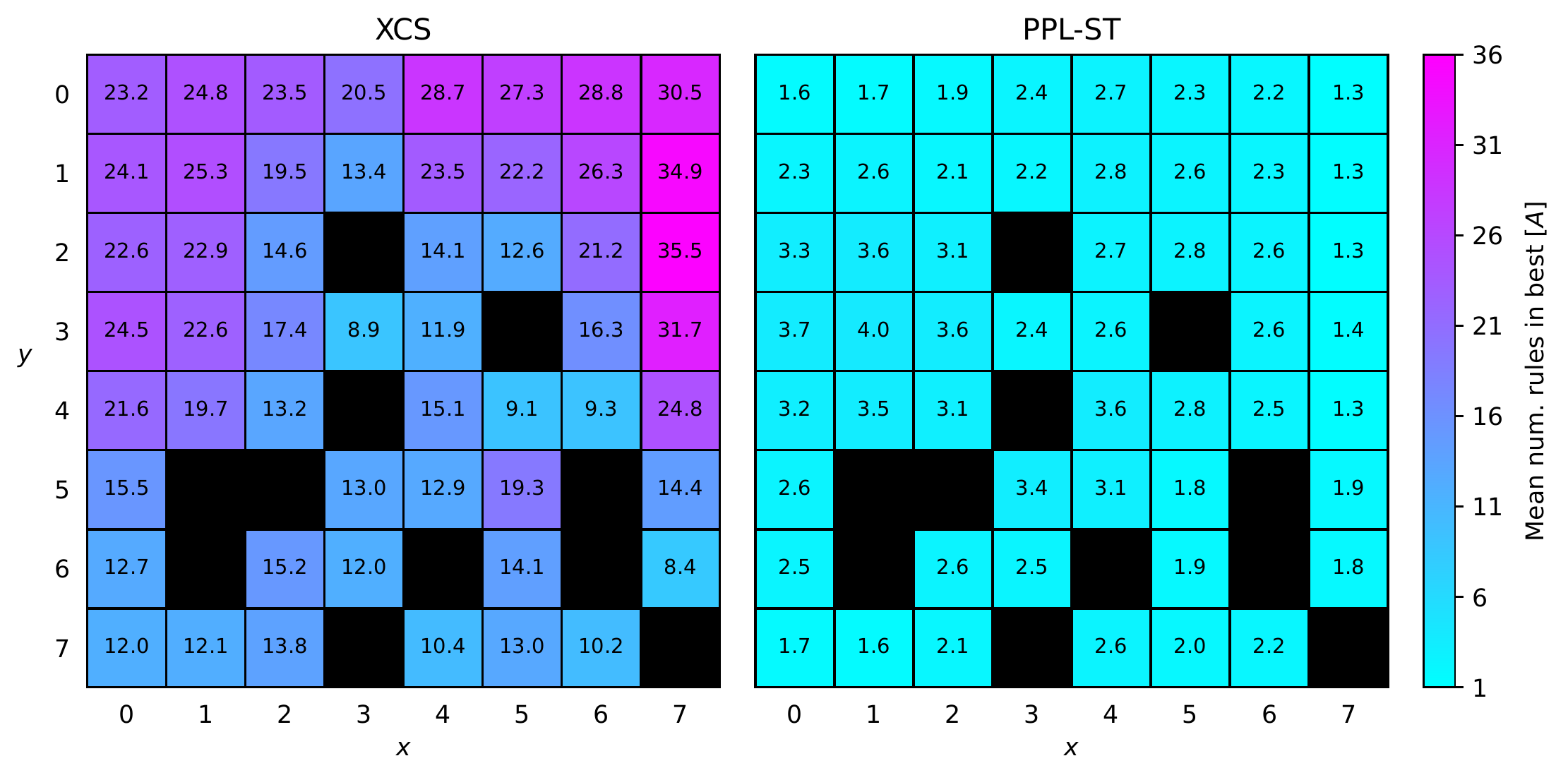}
    \caption{Local scale}
    \label{2fig:xcs_vs_pplst_bam_8_0.3_local}
\end{subfigure}
\hfill
\begin{subfigure}{0.425\textwidth}
    \includegraphics[width=\textwidth]{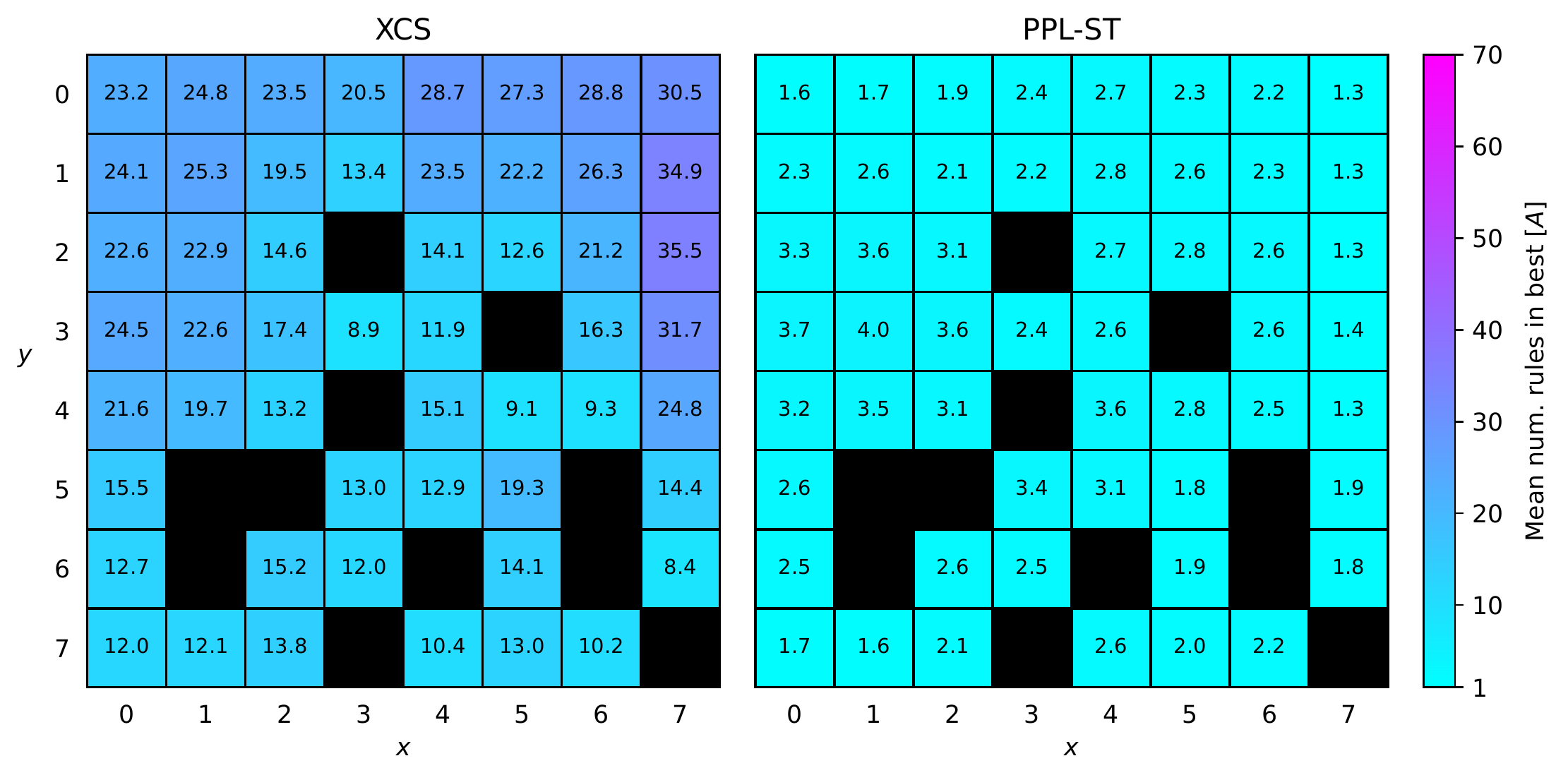}
    \caption{Global scale}
    \label{2fig:xcs_vs_pplst_bam_8_0.3_global}
\end{subfigure}
\caption{Mean best action set density of XCS and PPL-ST in $(8, 0.3)$ FL after 250 epochs.}
\label{fig:xcs_vs_pplst_bam_8_0.3_duo}
\end{figure}

\begin{figure}
\centering
\begin{subfigure}{0.425\textwidth}
    \includegraphics[width=\textwidth]{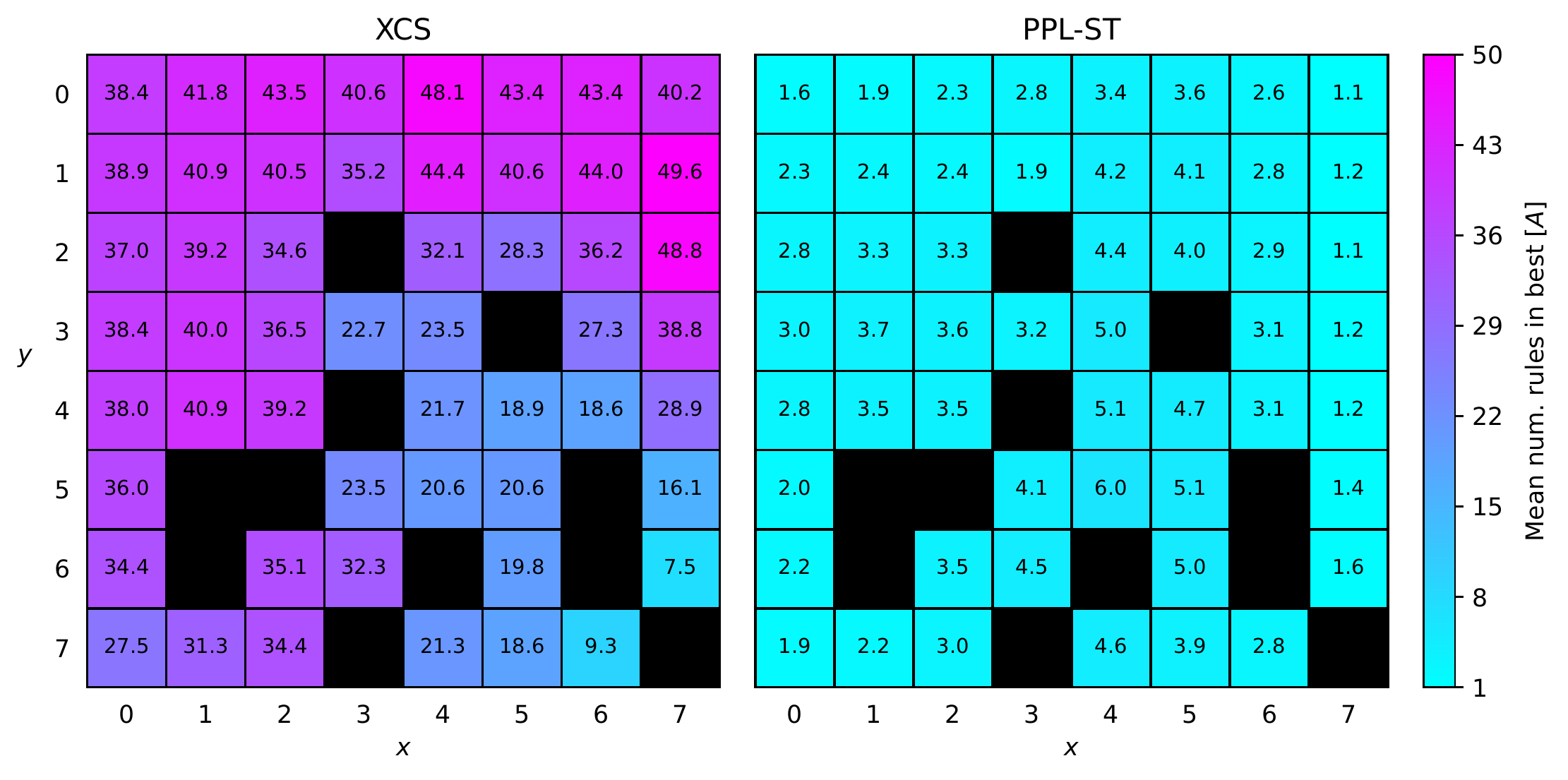}
    \caption{Local scale}
    \label{2fig:xcs_vs_pplst_bam_8_0.5_local}
\end{subfigure}
\hfill
\begin{subfigure}{0.425\textwidth}
    \includegraphics[width=\textwidth]{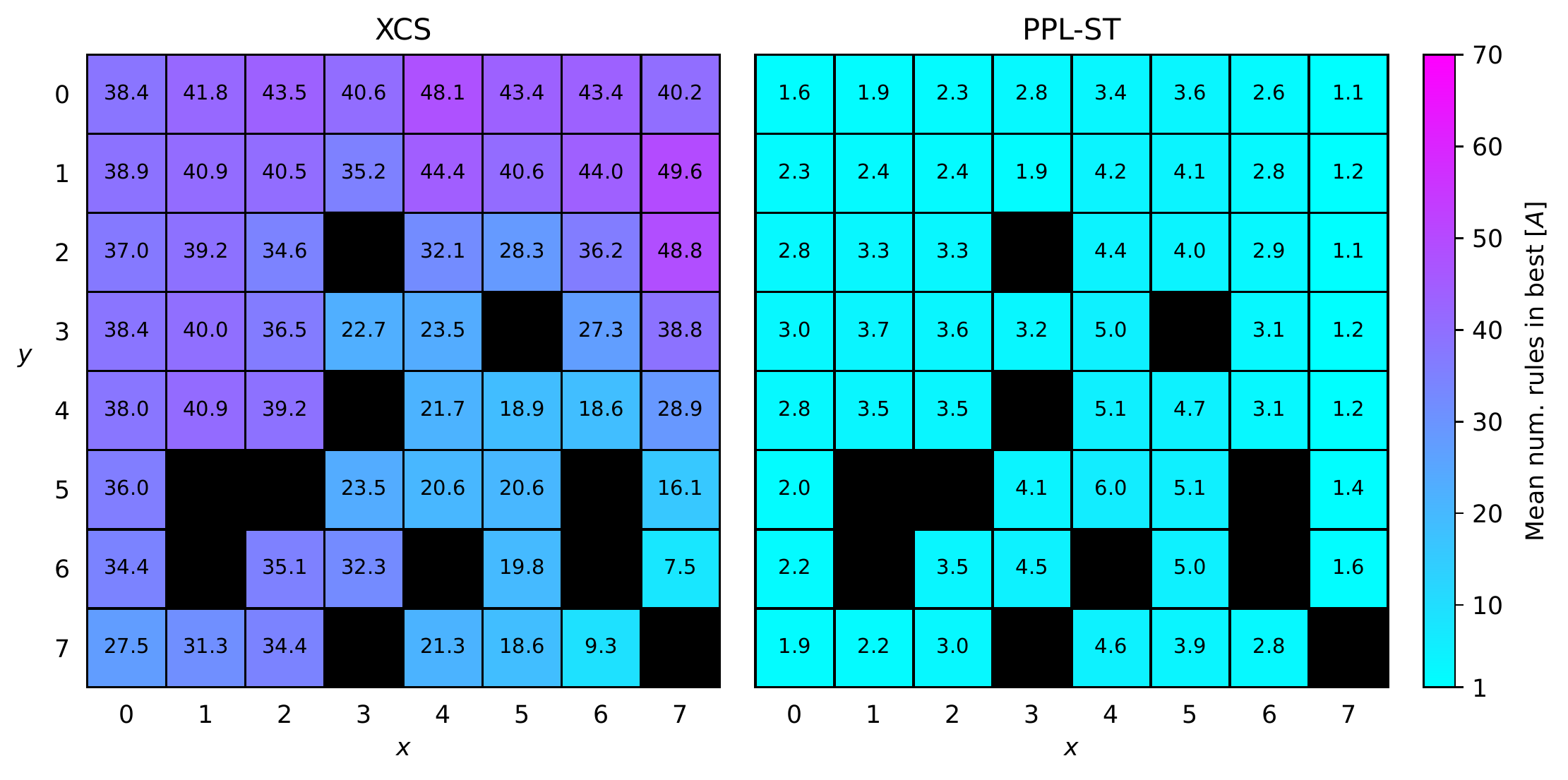}
    \caption{Global scale}
    \label{2fig:xcs_vs_pplst_bam_8_0.5_global}
\end{subfigure}
\caption{Mean best action set density of XCS and PPL-ST in $(8, 0.5)$ FL after 250 epochs.}
\label{fig:xcs_vs_pplst_bam_8_0.5_duo}
\end{figure}

\begin{figure}
\centering
\begin{subfigure}{0.425\textwidth}
    \includegraphics[width=\textwidth]{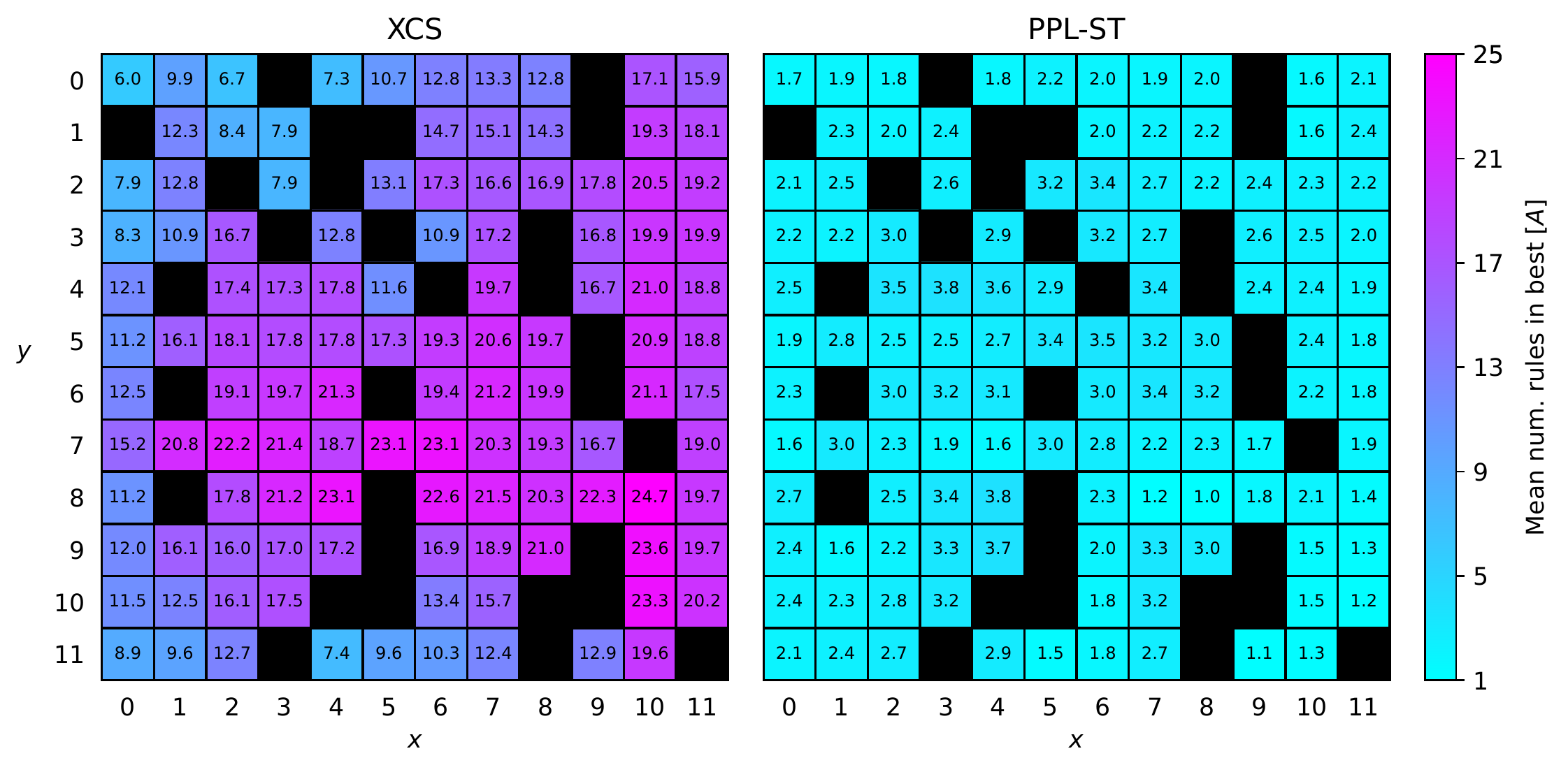}
    \caption{Local scale}
    \label{2fig:xcs_vs_pplst_bam_12_0_local}
\end{subfigure}
\hfill
\begin{subfigure}{0.425\textwidth}
    \includegraphics[width=\textwidth]{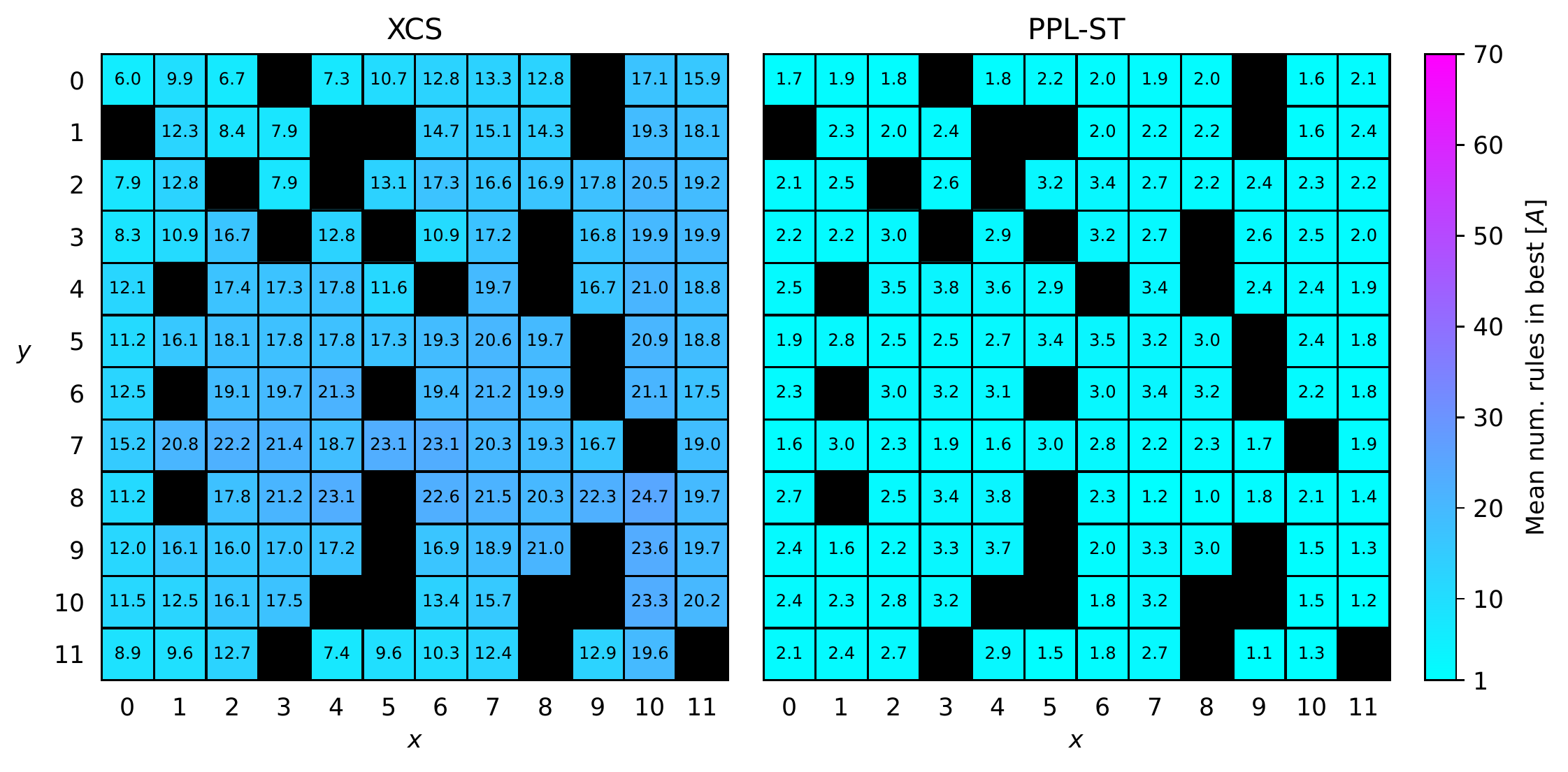}
    \caption{Global scale}
    \label{2fig:xcs_vs_pplst_bam_12_0_global}
\end{subfigure}
\caption{Mean best action set density of XCS and PPL-ST in $(12, 0)$ FL after 250 epochs.}
\label{fig:xcs_vs_pplst_bam_12_0_duo}
\end{figure}

\begin{figure}
\centering
\begin{subfigure}{0.425\textwidth}
    \includegraphics[width=\textwidth]{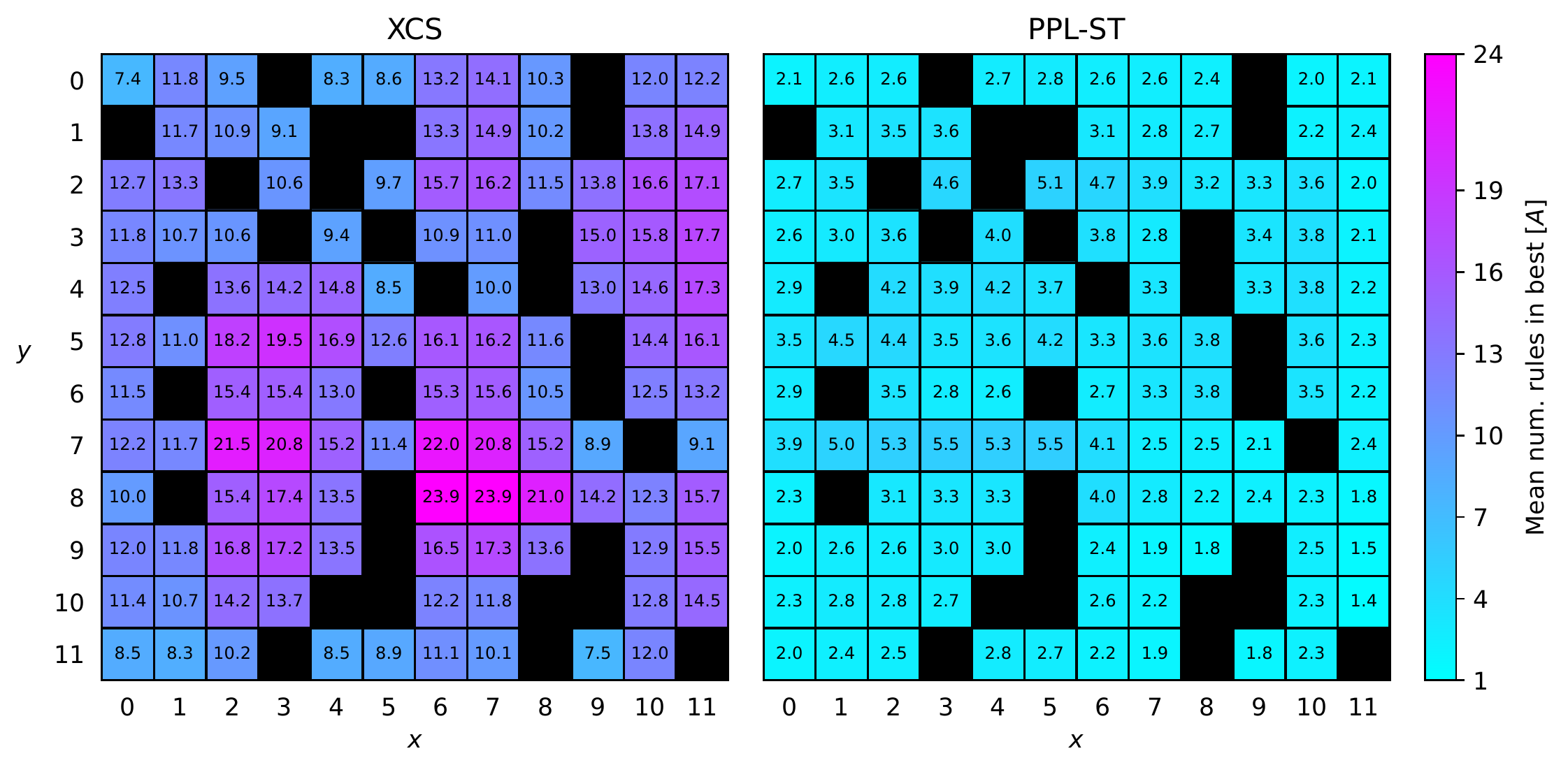}
    \caption{Local scale}
    \label{2fig:xcs_vs_pplst_bam_12_0.1_local}
\end{subfigure}
\hfill
\begin{subfigure}{0.425\textwidth}
    \includegraphics[width=\textwidth]{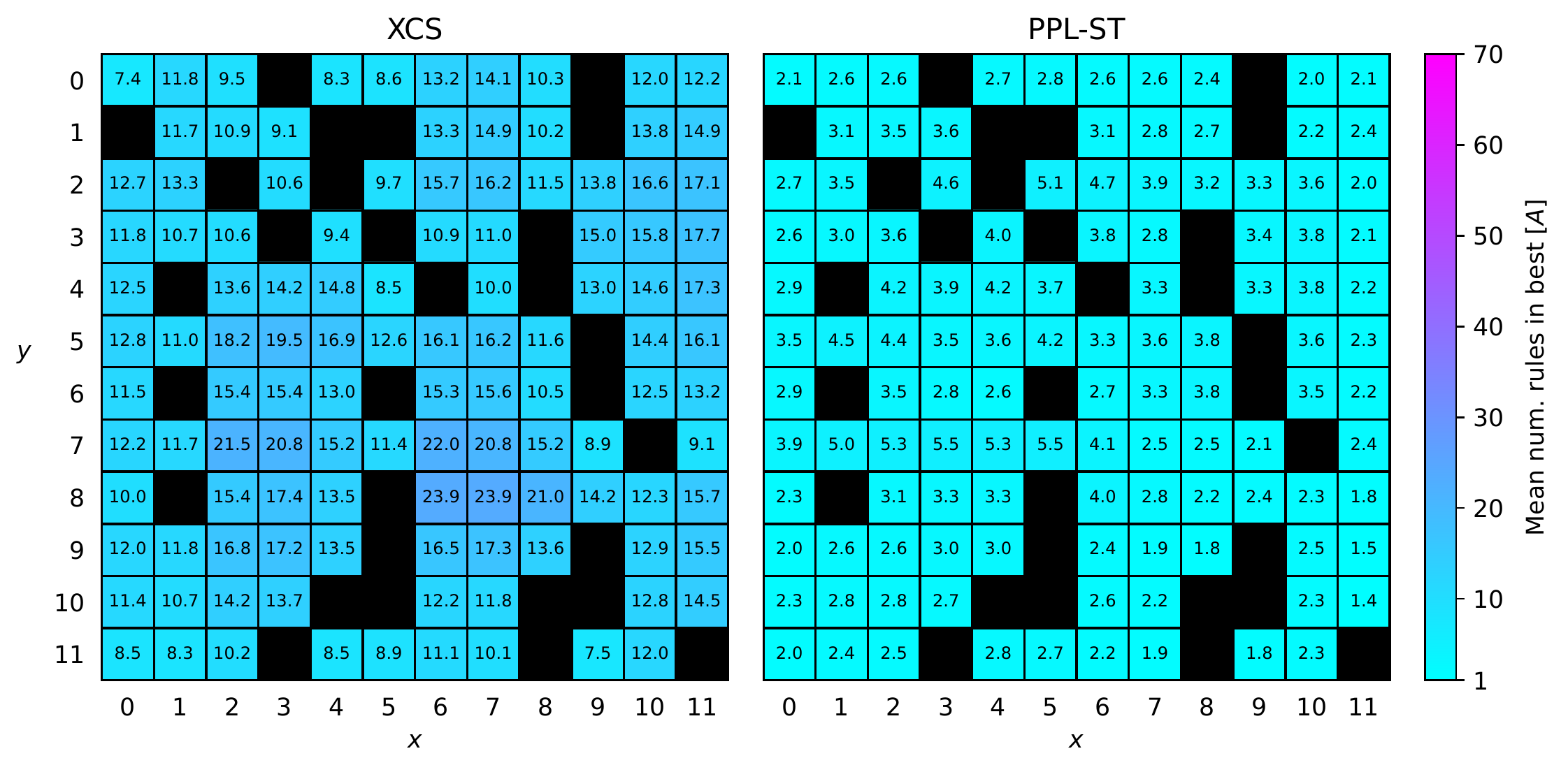}
    \caption{Global scale}
    \label{2fig:xcs_vs_pplst_bam_12_0.1_global}
\end{subfigure}
\caption{Mean best action set density of XCS and PPL-ST in $(12, 0.1)$ FL after 250 epochs.}
\label{fig:xcs_vs_pplst_bam_12_0.1_duo}
\end{figure}

\begin{figure}
\centering
\begin{subfigure}{0.425\textwidth}
    \includegraphics[width=\textwidth]{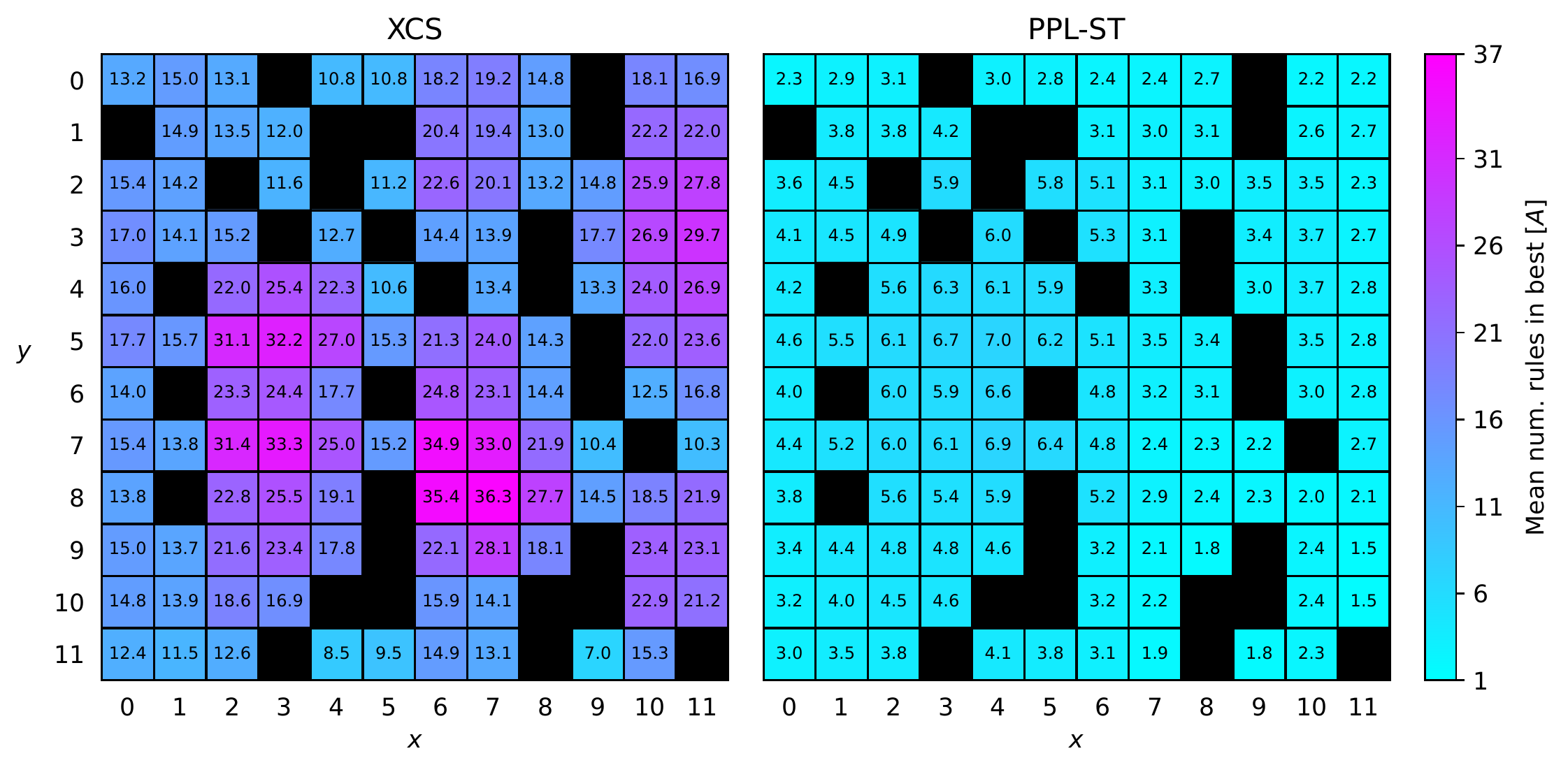}
    \caption{Local scale}
    \label{2fig:xcs_vs_pplst_bam_12_0.3_local}
\end{subfigure}
\hfill
\begin{subfigure}{0.425\textwidth}
    \includegraphics[width=\textwidth]{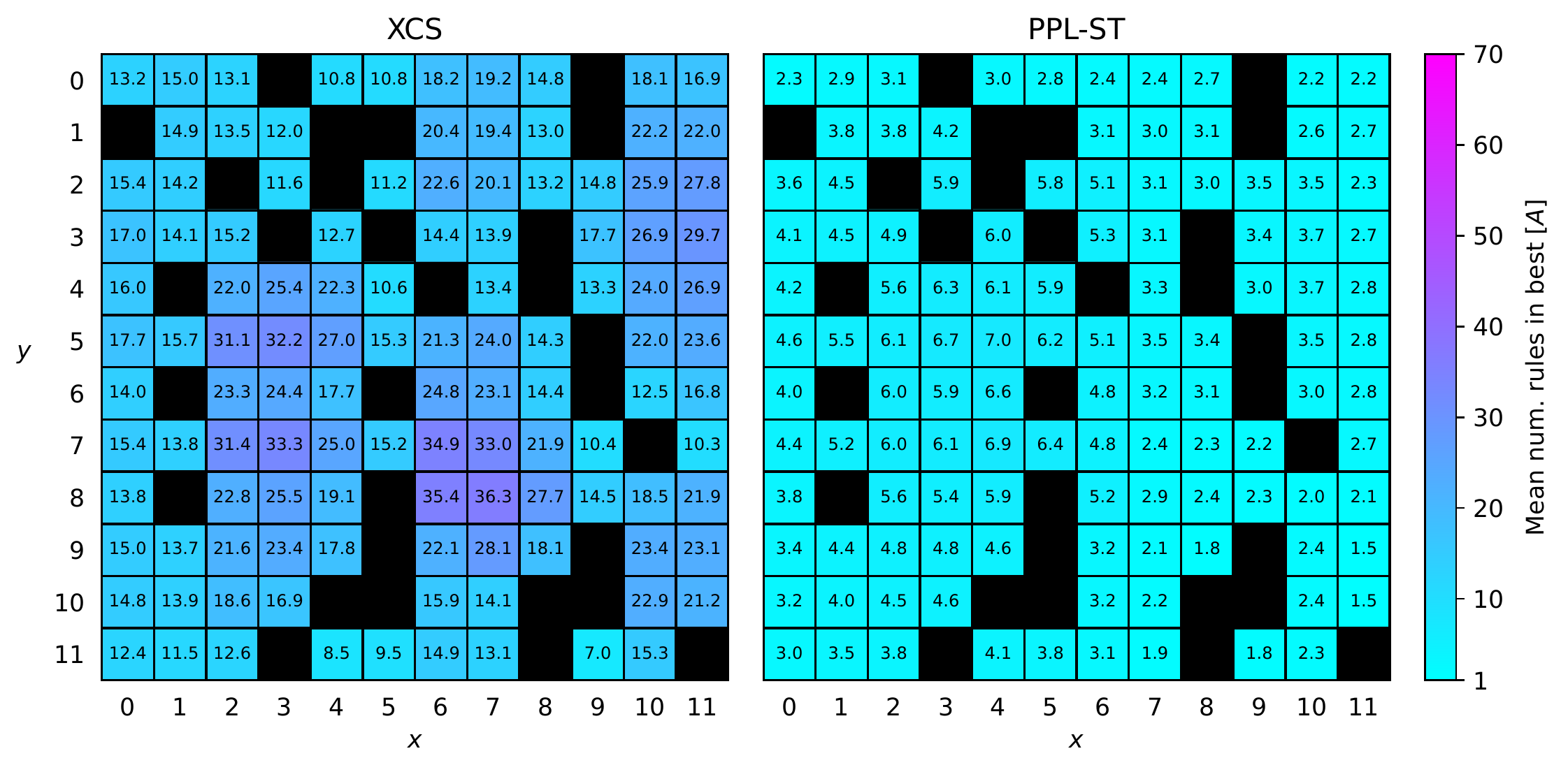}
    \caption{Global scale}
    \label{2fig:xcs_vs_pplst_bam_12_0.3_global}
\end{subfigure}
\caption{Mean best action set density of XCS and PPL-ST in $(12, 0.3)$ FL after 250 epochs.}
\label{fig:xcs_vs_pplst_bam_12_0.3_duo}
\end{figure}

\begin{figure}
\centering
\includegraphics[width=0.425\textwidth]{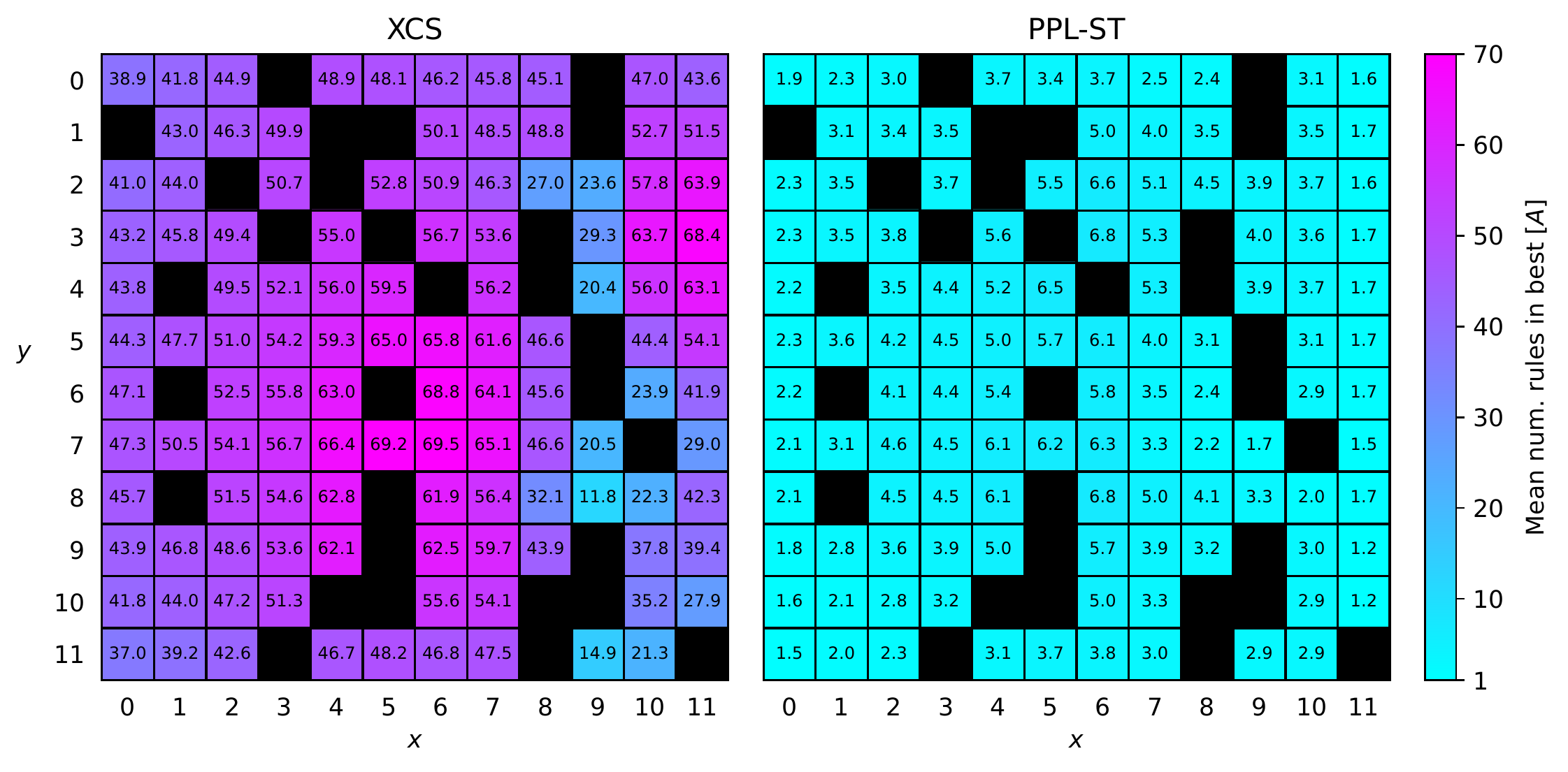}
\caption{Mean best action set density of XCS and PPL-ST in $(12, 0.5)$ FL after 250 epochs.}
\label{fig:xcs_vs_pplst_bam_12_0.5}
\end{figure}

Figures \ref{fig:xcs_vs_pplst_bam_4_0_duo} -- \ref{fig:xcs_vs_pplst_bam_12_0.5} show heatmaps of mean best action set density for XCS and PPL-ST across all FL environments after 250 epochs.
This is an extension of the results presented in Figure~\ref{fig:xcs_vs_pplst_action_set_heatmap} of the main paper, although note that Figure~\ref{fig:xcs_vs_pplst_action_set_heatmap} is equivalent to Figure~\ref{2fig:xcs_vs_pplst_bam_8_0.3_local} with the addition of relevant information from Table~\ref{tab:xcs_pplst_ruleset_sizes}.
Each of these figures (excluding \ref{fig:xcs_vs_pplst_bam_12_0.5}) shows two heatmaps, representing the same data on two different scales: local, which is environment-specific (left-hand subfigure) and global, which is shared across all environments (right-hand subfigure).
The minimum value on all scales was set to $1$.
The maximum value on the global scale was set based on the maximum value represented on any of the heatmaps; this occurring in the $(12, 0.5)$ environment.
Therefore there is only a single heatmap for the $(12, 0.5)$ environment in Figure~\ref{fig:xcs_vs_pplst_bam_12_0.5}, as the local and global scales for this environment are identical.
The intention of using both scales is to clearly visualise the patterns present in specific environments as well as across all environments.

% Please add the following required packages to your document preamble:
% \usepackage{multirow}
\begin{table}[h]
\centering
\caption{Ruleset size statistics for XCS and PPL-ST across all FL environments after 250 epochs. For both systems, $\vert \textit{R} \vert$ represents the size of the entire ruleset evolved by the system, and $\vert \textit{R}_{\text{BA}} \vert$ represents the size of the set of rules comprising the best action map.
For XCS both of these quantities are measured in units of number of macroclassifiers.
Overbars indicate sample means, computed over thirty trials.
Note that $\vert \textit{R} \vert$ for PPL-ST is not a sample mean because by definition this quantity is equivalent to the $\var{idvSize}$ hyperparameter, which is constant for each grid size (see experimental setup in Section~\ref{sec:results} of main paper).}
\label{tab:xcs_pplst_ruleset_sizes}
\renewcommand{\arraystretch}{1.35}
\begin{tabular}{ll|ll|ll|}
\cline{3-6}
                       &                      & \multicolumn{2}{l|}{\textbf{XCS}}   & \multicolumn{2}{l|}{\textbf{PPL-ST}}            \\ \cline{3-6} 
 &
   &
  \multicolumn{1}{l|}{$\bm{\overline{\vert \textbf{{\textit{R}}} \vert}}$} &
  $\bm{\overline{\vert \textbf{\textit{R}}_{\textbf{BA}} \vert}}$ &
  \multicolumn{1}{l|}{$\bm{\vert \textbf{\textit{R}} \vert}$} &
  $\bm{\overline{\vert \textbf{\textit{R}}_{\textbf{BA}} \vert}}$ \\ \hline
\multicolumn{1}{|l|}{\multirow{12}{*}{\textbf{\begin{tabular}[c]{@{}l@{}}FL\\ env.\end{tabular}}}} &
  $\bm{(4, 0)}$ &
  \multicolumn{1}{l|}{144.0} &
  63.8 &
  \multicolumn{1}{l|}{\multirow{4}{*}{7}} &
  6.0 \\ \cline{2-4} \cline{6-6} 
\multicolumn{1}{|l|}{} & $\bm{(4, 0.1)}$  & \multicolumn{1}{l|}{179.4}  & 68.9  & \multicolumn{1}{l|}{}                    & 6.7  \\ \cline{2-4} \cline{6-6} 
\multicolumn{1}{|l|}{} & $\bm{(4, 0.3)}$  & \multicolumn{1}{l|}{194.9}  & 83.4  & \multicolumn{1}{l|}{}                    & 6.5  \\ \cline{2-4} \cline{6-6} 
\multicolumn{1}{|l|}{} & $\bm{(4, 0.5)}$  & \multicolumn{1}{l|}{202.5}  & 108.2 & \multicolumn{1}{l|}{}                    & 6.5  \\ \cline{2-6} 
\multicolumn{1}{|l|}{} & $\bm{(8, 0)}$    & \multicolumn{1}{l|}{521.4}  & 261.9 & \multicolumn{1}{l|}{\multirow{4}{*}{21}} & 17.8 \\ \cline{2-4} \cline{6-6} 
\multicolumn{1}{|l|}{} & $\bm{(8, 0.1)}$  & \multicolumn{1}{l|}{594.6}  & 236.0 & \multicolumn{1}{l|}{}                    & 17.4 \\ \cline{2-4} \cline{6-6} 
\multicolumn{1}{|l|}{} & $\bm{(8, 0.3)}$  & \multicolumn{1}{l|}{615.7}  & 272.2 & \multicolumn{1}{l|}{}                    & 18.4 \\ \cline{2-4} \cline{6-6} 
\multicolumn{1}{|l|}{} & $\bm{(8, 0.5)}$  & \multicolumn{1}{l|}{591.7}  & 321.0 & \multicolumn{1}{l|}{}                    & 18.4 \\ \cline{2-6} 
\multicolumn{1}{|l|}{} & $\bm{(12, 0)}$   & \multicolumn{1}{l|}{1124.0} & 539.8 & \multicolumn{1}{l|}{\multirow{4}{*}{42}} & 33.1 \\ \cline{2-4} \cline{6-6} 
\multicolumn{1}{|l|}{} & $\bm{(12, 0.1)}$ & \multicolumn{1}{l|}{1264.2} & 527.3 & \multicolumn{1}{l|}{}                    & 34.5 \\ \cline{2-4} \cline{6-6} 
\multicolumn{1}{|l|}{} & $\bm{(12, 0.3)}$ & \multicolumn{1}{l|}{1458.9} & 697.6 & \multicolumn{1}{l|}{}                    & 35.2 \\ \cline{2-4} \cline{6-6} 
\multicolumn{1}{|l|}{} & $\bm{(12, 0.5)}$ & \multicolumn{1}{l|}{1303.1} & 790.4 & \multicolumn{1}{l|}{}                    & 34.1 \\ \hline
\end{tabular}
\end{table}

%% GS 4
\newpage

%% GS 8
\newpage

%% GS 12
\newpage

\end{document}